%% file: PaperForReview.tex
\documentclass[10pt,twocolumn,letterpaper]{article}

\usepackage{cvpr}              

\usepackage{graphicx}
\usepackage{amsmath}
\usepackage{amsfonts}
\usepackage{amssymb}
\usepackage{float}
\usepackage{booktabs}
\usepackage{bbold}
\usepackage[ruled]{algorithm2e}
\usepackage{algpseudocode}
\usepackage{fancyref}
\usepackage{amsthm}
\usepackage{macros}

\usepackage[pagebackref,breaklinks,colorlinks]{hyperref}
\usepackage{amsmath}
\usepackage{listings}
\usepackage{placeins}
\usepackage{indentfirst}

\DeclareMathOperator*{\argmin}{arg\,min}

\newcommand{\algname}{AutoSWAP}

\usepackage[capitalize]{cleveref}
\crefname{section}{Sec.}{Secs.}
\Crefname{section}{Section}{Sections}
\Crefname{table}{Table}{Tables}
\crefname{table}{Tab.}{Tabs.}

\def\cvprPaperID{7132} 
\def\confName{CVPR}
\def\confYear{2022}

\usepackage{xcolor}

\begin{document}
\title{Automatic Synthesis of Diverse Weak Supervision Sources for Behavior Analysis }


%

\author{
Albert Tseng \\ 
Nuro$^{*,\dagger}$
\and
Jennifer J. Sun \\
Caltech 
\and
Yisong Yue\\
Caltech, Argo AI$^*$
}

\maketitle

{
  \renewcommand{\thefootnote}%
    {\fnsymbol{footnote}}
  \footnotetext[1]{Work done while author was affiliated with Caltech.}
  \footnotetext[2]{Correspondence to \href{mailto:atseng@caltech.edu}{atseng@caltech.edu}.}
}

\begin{abstract}
Obtaining annotations for large training sets is expensive, especially in settings where domain knowledge is required, such as behavior analysis.
Weak supervision has been studied to reduce annotation costs by using weak labels from task-specific labeling functions (LFs) to augment ground truth labels. 
However, domain experts still need to hand-craft different LFs for different tasks, limiting scalability.
To reduce expert effort, we present {\algname}: a framework for automatically synthesizing data-efficient task-level LFs.
The key to our approach is to efficiently represent expert knowledge in a reusable domain-specific language and more general domain-level LFs, with which we use state-of-the-art program synthesis techniques and a small labeled dataset to generate task-level LFs.
Additionally, we propose a novel structural diversity cost that allows for efficient synthesis of diverse sets of LFs, further improving {\algname}'s performance.
We evaluate {\algname} in three behavior analysis domains and demonstrate that {\algname} outperforms existing approaches using only a fraction of the data. 
Our results suggest that {\algname} is an effective way to automatically generate LFs that can significantly reduce expert effort for behavior analysis.
\end{abstract}
\input{sec_introduction}

\input{sec_related}
\input{sec_methods}

\input{sec_experiments}
\input{sec_conclusion}
\section{Acknowledgements}
We thank Adith Swaminathan at Microsoft Research and Pietro Perona at Caltech for their feedback and  discussions regarding this work. We also thank Microsoft Research for compute resources. This work is partially supported by NSF Award \#1918839 (YY) and NSERC Award \#PGSD3-532647-2019 (JJS).
{\small
\bibliographystyle{ieee_fullname}
\bibliography{egbib}
}

\end{document}


\onecolumn

\title{Automatic Synthesis of Diverse Weak Supervision Sources for Behavior Analysis\\
Supplementary Materials}


%

\author{
Albert Tseng \\ 
Nuro$^*$
\and
Jennifer J. Sun \\
Caltech 
\and
Yisong Yue\\
Caltech, Argo AI$^*$
}

\maketitle

{
  \renewcommand{\thefootnote}%
    {\fnsymbol{footnote}}
  \footnotetext[1]{Work done while author was affiliated with Caltech.}
}

Here, we provide additional information about our datasets and experiments. The Supplementary Materials are organized into three sections: implementation details (Section~\ref{sec:implementation}), example learned programs (Section~\ref{sec:programs}), and additional results (Section~\ref{sec:add_results}).

\input{supp_implementation.tex}
\input{supp_programs.tex}
\input{supp_add_results.tex}

{\small
\bibliographystyle{ieee_fullname}
\bibliography{egbib}
}

%% file: sec_introduction.tex
\section{Introduction}
\label{sec:intro}
In recent years, machine learning has enabled the study of large-scale datasets in many behavior analysis domains, such as neuroscience~\cite{segalin2020mouse,sun2021multi}, sports analytics~\cite{zhan2018generating,tuyls2021game}, and motion forecasting~\cite{chang2019argoverse}.
However, obtaining labeled data to train models can be difficult and costly, especially when domain expertise is required for annotation, such as for many behavior analysis tasks~\cite{segalin2020mouse}.  
One way to reduce annotation cost is through weak supervision, which uses noisy, task-level heuristic ``labeling functions'' (LFs) to weakly label data.
LFs for a specific task (task-level LFs) are supplied by domain experts, and are applied to obtain a set of weak labels.
Weakly labeled data can then be used in downstream settings, such as active learning~\cite{biegel2021active} and self-training~\cite{karamanolakis2021self}.

While weak supervision has worked well in a wide range of settings ~\cite{ratner2016data,biegel2021active,dunnmon2019}, it has not been well-explored for behavior analysis tasks.
For one, the requirement that LFs must provide \textit{labels} and not, for example, \textit{features} prevents more general domain knowledge from being used~\cite{ratner_metal} (e.g. the behavioral features in~\cite{eyjolfsdottir2014detecting,segalin2020mouse}). 
Furthermore, new LFs must be hand-crafted by domain experts for new tasks (such as new behaviors to study), limiting the scalability of manual weak supervision~\cite{varma2018snuba}.
To address these challenges, we study efficient domain knowledge representations and develop automated weak supervision methods towards reducing annotation bottlenecks in behavior analysis settings.

\begin{figure}[t]
\includegraphics[width=\linewidth]{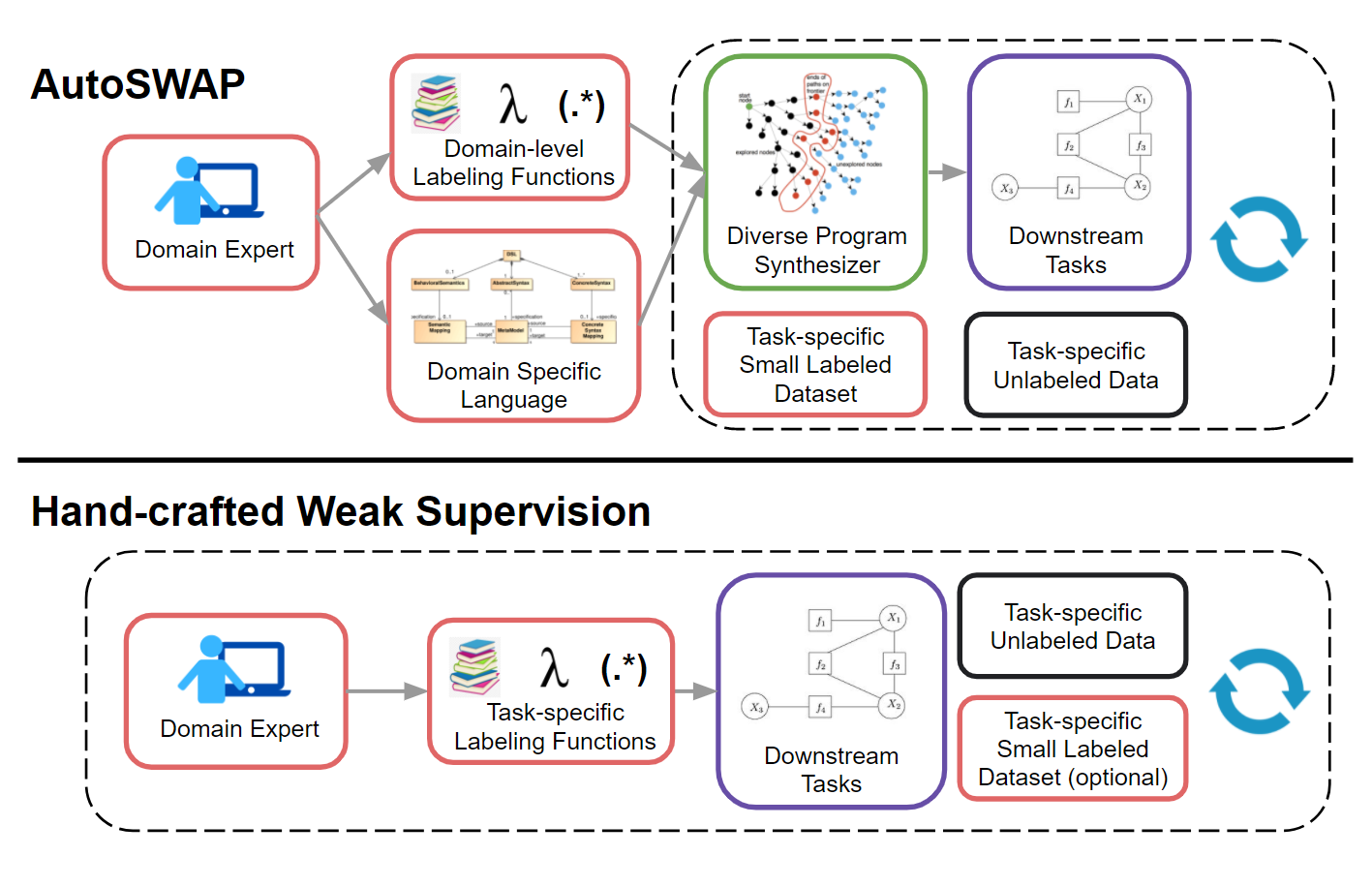}
\vspace{-0.2in}
\caption{We present {\algname}, a framework for automatically synthesizing diverse sets of \textit{task-level} labeling functions (LFs) with a small labeled dataset and domain knowledge encoded in \textit{domain-level} LFs and a DSL.
{\algname} significantly reduces labeler effort by automating LF generation.}
\label{fig:teaser}
\end{figure}

\textbf{Our Approach.}
We propose {\algname} (\textbf{Auto}matic \textbf{S}ynthesized \textbf{W}e\textbf{A}k Su\textbf{P}ervision), a data-efficient framework for automatically generating task-level LFs using a novel diverse program synthesis formulation.
As depicted in Figure \ref{fig:teaser}, experts provide a domain-specific language (DSL) and domain-level LFs (LFs specific to a domain of tasks) for a given domain, such as mouse behaviors or vehicle motion planning. 
For each task to be studied in that domain, experts provide a small labeled dataset to specify the task, and AutoSWAP returns a set of structurally diverse task-level LFs that can be used in weakly supervised frameworks.
The domain-level LFs (Figure \ref{fig:example_lfs}) provide fine-grained, label-space agnostic ``atomic instructions,'' while the DSL contains abstract structural domain knowledge for composing the more general domain-level LFs into task-level LFs (Figure \ref{fig:prog2tree}).
The novel diversity cost enables {\algname} to generate structurally diverse LFs, which we and others empirically show outperform structurally homogeneous LFs in downstream tasks~\cite{varma2018snuba}.
 
To the best of our knowledge, we are the first to demonstrate the effectiveness of program synthesis for automated LF generation. 
Existing works for generating LFs include iteratively selecting LFs by repeatedly querying experts for feedback~\cite{boecking2020interactive} and training exponentially many simple heuristics models~\cite{varma2018snuba}, which have limitations in scalability and tractability. 
In contrast, our approach represents domain knowledge in a DSL and domain-level LFs, which can then be used to automatically synthesize LFs for arbitrary tasks in a domain with our diverse program synthesizer. 

We evaluate our approach in three behavior analysis domains with both sequential and nonsequential data: mouse~\cite{sun2021multi}, fly~\cite{eyjolfsdottir2014detecting}, and basketball player~\cite{yue2014learning} behaviors. 
In these domains, data collection is expensive and new tasks frequently emerge, highlighting the importance of scalability.
The datasets we use are based on agent trajectories, which provide low-dimensional inputs for easily creating domain-level LFs.
We show that with existing expert defined domain-level LFs from \cite{segalin2020mouse,eyjolfsdottir2014detecting} and a simple DSL, {\algname} is capable of synthesizing high quality LFs with very little labeled data.
These LFs outperform LFs from existing automatic weak supervision methods~\cite{varma2018snuba} and offer a data efficient approach to reducing domain expert effort. 

To summarize, our contributions are: 
\begin{itemize}
    \item We propose {\algname}, which combines program synthesis with weak supervision to scalably and efficiently generate labeling functions.  
    \item We propose a novel program-structural diversity cost that enables \algname~to directly synthesize diverse sets of labeling functions, which we empirically show are more data efficient than purely optimal sets.
    \item We evaluate {\algname} in multiple behavior analysis domains and downstream tasks, and show that {\algname} is capable of significantly improving data efficiency and reducing expert cost.
\end{itemize}

Our implementation of AutoSWAP can be found at \href{https://github.com/autoswap/autoswap_cvpr_2022}{https://github.com/autoswap/autoswap\_cvpr\_2022}.

%% file: sec_related.tex
\section{Related Work}~\label{sec:related_works}

\textbf{Behavior Analysis}.
In many domains, such as behavioral neuroscience~\cite{segalin2020mouse,luxem2020identifying}, sports analytics~\cite{yue2014learning,zhan2018generating}, and traffic modeling~\cite{colyar2007us}, agent pose and location trajectory data is used for behavior analysis.
This data is usually extracted from recorded videos using detectors and pose estimators~\cite{eyjolfsdottir2014detecting,segalin2020mouse};
for example, we use trajectories from \cite{segalin2020mouse}, ~\cite{eyjolfsdottir2014detecting}, and StatsPerform for our mouse, fly, and basketball datasets, respectively.

To accurately analyze this data for complex behaviors, frame-level behavior labels from domain experts are usually needed.
However, annotating large datasets is time-consuming and monotonous~\cite{anderson2014toward}, motivating methods for label-efficient modeling.
For example, self-supervised learning \cite{sun2020task} and unsupervised behavior discovery methods \cite{berman2014mapping,luxem2020identifying,calhoun2019unsupervised} aim to learn efficient behavior representations and discover new behaviors, respectively.
Our work is complementary to these methods in that this is not a comparison between weak supervision and self-supervision.
Rather, we evaluate the merits of our synthesized LFs in the context of weak supervision for learning expert-defined behaviors.

\textbf{Weak Supervision}.
Weak supervision with LFs was introduced in the context of data programming~\cite{ratner2016data}.
Since then, LFs have been applied in a variety of settings, including for active learning~\cite{nashaat2018hybridization,biegel2021active} and self-training~\cite{karamanolakis2021self} tasks. 
Our work is complementary to these works in that we automatically learn LFs that can be used as inputs to existing weakly supervised frameworks.
We note that we are not the first to propose learning LFs from a small amount of training data. 
For example, IWS iteratively proposes rules and queries domain experts in a large-scale feedback loop \cite{boecking2020interactive}.
More similar to our work, SNUBA~\cite{varma2018snuba} trains heuristics models, but does so without domain knowledge and has runtime exponential in the number of features. 
To the best of our knowledge, we are the first to apply program synthesis to this problem, and our framework outperforms existing model-based methods for learning LFs. 


\textbf{Program Synthesis}.
Traditionally, programming by example has been used to synthesize programs from a DSL that respect hard constraints on input/output examples~\cite{solar2006combinatorial,feser2015synthesizing}.
In recent years, a growing number of works have studied synthesizing programs with soft constraints, such as minimizing a loss function ~\cite{shah2020learning,ellis2015unsupervised,parisotto2016neuro,valkov2018houdini}. 
This relaxed form of program synthesis has been applied to a number of different domains including web information extraction~\cite{chen2021web}, image structure analysis~\cite{ellis2017learning}, and learning interpretable agent policies~\cite{verma2018programmatically}.
Of these works, algorithms that learn differentiable programs, such as~\cite{shah2020learning}, have shown great promise in being able to efficiently and simultaneously optimize program architectures and parameters. 
Here, we use concepts from differentiable program synthesis algorithms to synthesize diverse sets of LFs.

%% file: sec_methods.tex
\section{Methods}~\label{sec:methods}
We introduce {\algname}, a framework for automatically generating diverse sets of task-level LFs.
In our framework, domain experts provide a set of domain-level LFs and a DSL of useful relations.
For each task to be studied, specified with a small labeled dataset, task-level LFs are automatically generated by the {\algname} diverse program synthesizer.
These LFs can then be used in downstream applications involving weak supervision.
In the following sections, we provide a background of key components in {\algname} (Section \ref{sec:methods_bg}), detail the framework (Section \ref{sec:methods_main_all}), and describe example downstream applications (Section \ref{sec:downstream}). 

\subsection{Background} \label{sec:methods_bg}
\definecolor{light-gray}{gray}{0.95}
\begin{figure}
\centering
\lstset{
  basicstyle=\footnotesize,
  frame=tb,
  language=Python,
  backgroundcolor=\color{light-gray},
  numbers=none
}
\begin{lstlisting}
# lambda_1 - whether fly is attacking target
def is_attacking(fly, tgt):
    f2t_angle = atan((tgt.y - fly.y) / (tgt.x - fly.x))
    rel_angle = |fly.abs_angle - f2t_angle|
    return fly.speed > 2 and rel_angle < 0.1

# lambda_2 - ratio of fly wingspan
def wing_ratio(fly, tgt):
    return quantize(fly.wing_x / fly.wing_y, 4)

# lambda_3 - fly speed relative to target speed
def relative_speed(fly, tgt):
    return |fly.speed| / |tgt.speed|
\end{lstlisting}
\vspace{-0.1in}
\caption{Domain experts provide domain-level labeling functions, such as the ones above for the fly domain. Some domain-level LFs ($\lambda_1$, $\lambda_2$) label for specific tasks (and would be considered task-level LFs on their own), while others ($\lambda_3$) return features.}
\label{fig:example_lfs}
\end{figure}

\textbf{Domain-level Labeling Functions}.
In weak supervision, users provide a set of \textit{task-level} hand-crafted heuristics called labeling functions (LFs).
LFs can be noisy and abstain from labeling, but LFs must output in downstream task's label space $\mathcal Y$.
We relax this requirement in {\algname} by allowing domain experts to provide \textit{domain-level} LFs (Figure \ref{fig:example_lfs}).
These LFs do not have to output in $\mathcal Y$, which reduces LF creation overhead and allows for more expressive LFs.
This also allows us to reuse LFs across multiple tasks within the same domain, aiding scalability.

\textbf{Domain Specific Languages}.
Domain specific languages (DSLs) define the allowable submodules and structures in synthesized programs, and are a key component of program synthesis algorithms.
Many recent works have adopted purely functional DSLs~\cite{shah2020learning}, where DSL items are functions that output to the input space of other DSL items or the final output space.
In {\algname}, domain experts provide a purely functional DSL with program structures that may be useful in generated LFs.
We show empirically that even using a very simple DSL in {\algname} can result in significant reductions in expert effort.



\textbf{Differentiable Program Synthesis via Neural Completions and Guided Search}.
Our program synthesis formulation is based on NEAR, which finds $\epsilon$-optimal differentiable programs using admissible search heuristics~\cite{harris, shah2020learning}.
While NEAR is one instantiation of {\algname}, our diverse synthesis formulation (Section \ref{sec:methods_main_all}) is theoretically compatible with any search-based synthesizer.
Here, the DSL $\mathcal D$ is a context-free grammar with differentiable variables.
Programs are defined by a program architecture $\alpha$ in the context-free language of $\mathcal D, \mbox{CFL}_{\mathcal D}$, and a set of real parameters $\theta$, and are denoted by $\Sem{\alpha}(x, \theta): \mathcal{X} \rightarrow \mathcal{Y}$.
Synthesizing a program that is optimal w.r.t. a cost function $F$ and dataset $(X, Y) \in (\mathcal X, \mathcal Y)$ is equivalent to
\begin{equation}
\label{eqn:optimal_diff_prog}
(\alpha^*, \theta^*) = \argmin_{\alpha, \theta} F(\Sem{\alpha}(X, \theta), Y).
\end{equation}

To find $(\alpha^*, \theta^*)$, we search over $\mbox{CFL}_{\mathcal D}$.
This search space is a tree $\mathcal G$, where the root node is an empty architecture, interior nodes are incomplete architectures (architectures with unknown components), and leaf nodes are complete architectures.
Edges in $\mathcal G$ represent single productions from $\mathcal D$ between two architectures. 
We bound the search tree by limiting the search depth to $m$ and ``completing'' incomplete architectures by substituting unknown components with neural networks (``neural completions'').

Since neural completions are differentiable, the minimum cost-to-go (CTG) w.r.t.~$F$ of a neural completion can be computed by optimizing the neural completion's parameters.
Furthermore, this minimum CTG of a neural completion is an $\epsilon$-admissible heuristic\cite{harris} for the true CTG of the corresponding incomplete architecture (proof in \cite{shah2020learning}).
This allows us to use informed search algorithms on $\mathcal G$ to find $\epsilon$-optimal solutions to Equation \ref{eqn:optimal_diff_prog}.

\subsection{{\algname}}
\label{sec:methods_main_all}
\textbf{Synthesizing Diverse Sets of Programs}.
Diverse sets of LFs have been shown to improve data efficiency relative to purely optimal sets in downstream applications of weak supervision~\cite{varma2018snuba}.
This is partly due to diverse sets having improved label coverage (fewer data points where all LFs abstain) \cite{varma2018snuba}, and from having more learning signals for the downstream model \cite{sun2018dt}.
The program synthesizer in Section \ref{sec:methods_bg} can be run repeatedly to obtain a set of purely optimal LFs, but there is no guarantee that the set will be diverse. 
Here, we introduce a structural diversity cost and admissible heuristic that allows for direct synthesis of diverse sets of programs using informed search algorithms. 
We empirically show that using the diversity cost improves performance, corroborating \cite{varma2018snuba}'s observations.

\begin{figure}
\centering
\includegraphics[width=\linewidth]{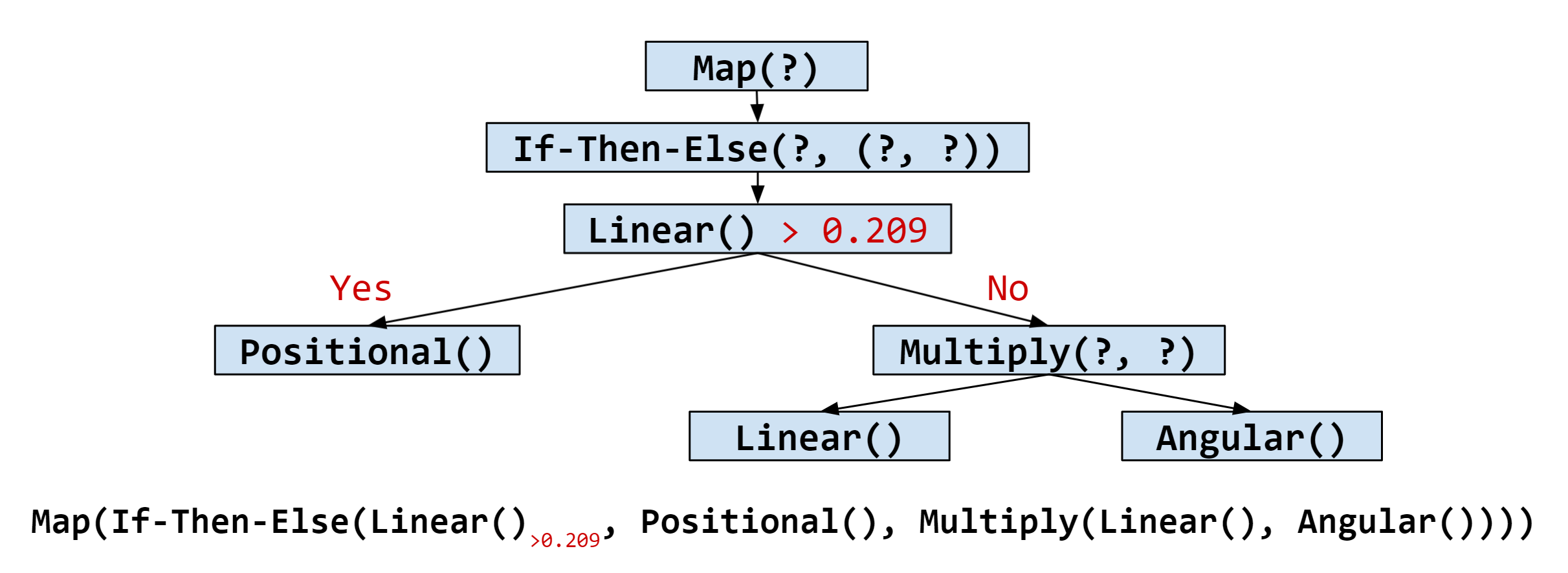}
\vspace{-0.2in}
\caption{A complete program and its tree representation. Each `\texttt{?}' represents one child node function. The depicted program is an actual {\algname} LF for the ``lunge vs. no behavior'' in the Fly domain. The program can be interpreted as \textit{``If the linear speed between the flies is small, classify the angular domain-level LFs of the flies. Otherwise, classify the product of transformations of the linear speed and positional domain-level LFs.''} Note the parameters (red) are not included in the structural diversity cost.}
\label{fig:prog2tree}
\end{figure}

Consider a complete program $P$, which is a composition of variables in $\mathcal D$. 
By construction of $\mathcal G$, we can convert $P$ to a tree $T_P$ where each node is a variable in $P$ and a node's children are its input variables (Figure \ref{fig:prog2tree}).
Then, given a set of complete programs $\mathcal P$ and a complete program $P$, we define the structural cost $C_{P, \mathcal P}$ of $P$ relative to $\mathcal P$ as:
\begin{equation}
\label{eqn:struct_cost}
\frac{1}{C_{P, \mathcal P}} = q\Big(\frac{1}{\|\mathcal P\|} \sum_{P' \in \mathcal P} \ZSS(T_P, T_{P'}) \Big),
\end{equation}
where $q:\mathbb{R} \to \mathbb{R}$ is a user defined monotonically increasing function and $\ZSS$ is the Zhang-Shasha tree edit distance (TED)~\cite{zss}.
Essentially, programs with a higher average TED to the elements of $\mathcal P$ incur a lower diversity cost.

Since this structural cost is not defined for incomplete programs or neural completions, $C_{P, \mathcal P}$ cannot be used in informed search algorithms.
However, the following admissible heuristic $H_{P_I, \mathcal P}$ for incomplete programs $P_I$ allows us to create a set of diverse programs by iteratively synthesizing programs and adding them to $\mathcal P$.

\newtheorem{theorem}{Theorem}[section]
\newtheorem{lemma}[theorem]{Lemma}
\begin{lemma}
Let $P_I$ be an incomplete program and $T_{P_I}$ be the tree of its known variables.
$T_{P_I}$ is guaranteed to exist by construction of $\mathcal G$.
Define $H_{P_I, \mathcal P}$ as:
\begin{align*}
U_{P_I, P'} &= m - \|P_I\| + \ZSS(T_{P_I}, T_{P'}),\\
\frac{1}{H_{P_I, \mathcal P}} &= q\Big(\frac{1}{\|P\|} \sum_{P' \in \mathcal P} U_{P_I, P'} \Big),
\end{align*}
where $\|P_I\|$ is the number of known variables in $P_I$.
$H_{P_I, \mathcal P}$ is an admissible heuristic for the CTG from $P_I$ in $\mathcal G$.
\end{lemma}

\begin{proof}
Consider $U_{P_I, P'}$. 
$m - \|P_I\|$ is an upper bound on the TED between $T_{P_I}$ and the tree of any complete descendant $P^*$ of $P_I$ in $\mathcal G$.
From the triangle inequality, 
\begin{align*}
U_{P_I, P'} &= m - \|P_I\| + \ZSS(T_{P_I}, T_{P'}) \\
&\ge \ZSS(T_{P_I}, T_{P^*}) + \ZSS(T_{P_I}, T_{P'}) \\
&\ge \ZSS(T_{P^*}, T_{P'}).
\end{align*}
Then, as TEDs are nonnegative, $m \ge \|P_I\|$, and $q$ is nondecreasing, $H_{P_I, \mathcal P} \le C_{P^*, \mathcal P}$.
Thus, $H$ a admissible heuristic for the structural CTG from $P_I$.
\end{proof}

\textbf{{\algname} Framework}.
{\algname} uses program synthesis to automate significant parts of the weak supervision pipeline and reduce domain expert effort.
Domain experts provide a set of domain-level LFs $\Lambda_m = \{\lambda_i: \mathcal X \to \mathcal Y_i\}$, a purely functional DSL $\mathcal D$, and a small labeled dataset $(X, Y) \in (\mathcal X, \mathcal Y)$ to specify tasks within the domain.
In order to use $\Lambda_m$ when synthesizing programs with $\mathcal D$, all $\lambda_i$ must be added to $\mathcal D$.
This can be done either by implementing each $\lambda_i$ with operations from $\mathcal D$, or precomputing and selecting $\Lambda_m(X)$ as input features in $\mathcal D$; we do the latter in our experiments. 
With $\mathcal D$, {\algname} runs the diverse program synthesis algorithm $n$ times to generate a set $\Lambda$ of $n$ LFs.
$\Lambda$ can then be used in downstream tasks, such as in weak supervision label models to generate weak labels.
See Algorithm \ref{alg:main_alg} for a detailed description of {\algname}.

\begin{algorithm}[t]
\KwIn {$\Lambda_m$, $\mathcal D$, labeled dataset $D_L$, \# LFs $n$}
\KwOut{task-level LFs $\Lambda$}
$\mathcal D \gets $ Combine $\Lambda_M$ and $\mathcal D$\\
$\mathcal P \gets \emptyset$\\
\While {$\|\mathcal P\| \le n$} {
    Synthesize $P$ with $\mathcal D, D_L, \mathcal P$ \\
    $\mathcal P \gets \mathcal P \cup \{P\}$
}
$\Lambda \gets \mathcal P$, return $\Lambda$.
\caption{{\algname}.} \label{alg:main_alg}
\end{algorithm}

\subsection{Downstream Tasks}
We describe two downstream tasks in which weak labels can be used. These examples, which our experiments are based on, are just a subset of the many weakly supervised learning frameworks in existence such as ASTRA~\cite{karamanolakis2021self}.
\label{sec:downstream}

\textbf{Active Learning}.
Active learning is a paradigm where the learning algorithm can selectively query for new data to be labeled.
Here, we use labels from task-level LFs as additional features for a downstream classifier. 
The downstream classifier's predictions are used to select data for labeling.
To evaluate generated LFs in active learning settings, we consider the performance of downstream classifiers at multiple data amounts.
Given a sorted list $A$ of data amounts, at each amount we generate new LFs, train a downstream classifier, and select data points for labeling to form the next batch.
An exact description of our active learning setup for {\algname} can be found in Algorithm \ref{alg:active_learning}.

\begin{algorithm}[t]
\KwIn {$\Lambda_m, \mathcal D, n$, unlabeled $X_U$, $A$.}
Sort $A$ in increasing order.\\
Randomly select $A_1$ points $X_L$ from $X_U$. \\
$X_U \gets X_U \setminus X_L$ \\
$Y_L \gets$ Obtain labels for $X_L$.\\
\For{$i = 1, ..., \|A\| - 1$}{
$\Lambda_i \gets$ AutoSWAP($\Lambda_m, \mathcal D, (X_L, Y_L), n$).\\
$X'_L \gets \begin{bmatrix} X_L & \Lambda_i(X_L)\end{bmatrix}$\\
Train downstream classifier $C_i$ with $(X'_L, Y_L)$.\\
Select $A_{i+1} - A_i$ points $X'_L$ using max entropy uncertainty sampling.\\
$X_U \gets X_U \setminus X'_L$ \\
$X_L \gets X_L \cup X'_L$ \\
$Y_L \gets Y_L \cup \{\mbox{Obtain labels for } X'_L\}$
}
\caption{{\algname} for Active Learning.} \label{alg:active_learning}
\end{algorithm}

\textbf{Weak Supervision}.
Weak supervision frameworks generally depend on a generative label model weakly label unlabeled samples.
Using no ground truth labels, the generative model produces probabilistic estimates (``weak labels'') for the true labels $Y_U$ of an unlabeled set $X_U$ by modeling the LF outputs $\Lambda(X_U)$.
Weakly labeled data can then be used to augment labeled datasets in downstream tasks.

To evaluate {\algname} in weak supervision settings, we start with a small labeled dataset $D_L$ and a list of unlabeled data amounts $A$.
LFs are generated using the small labeled dataset and abstain using the method in \cite{varma2018snuba}.
Then, weak labels are generated from these LFs for all unlabeled data using the generative model.
For each data amount $A_i \in A$, a random set $D_{PL}$ of $A_i$ weakly labeled data points is selected and the performance of a downstream classifier is measured using the training set $D_L \cup D_{PL}$.
An exact description of our weak supervision setup is in Algorithm \ref{alg:weak_sup}.
\begin{algorithm}[t]
\KwIn {$\Lambda_m, \mathcal D, n$, Labeled $(X_L, Y_L)$, Unlabeled $X_U$, $A$.}
$\Lambda \gets$ AutoSWAP($\Lambda_m, \mathcal D, (X_L, Y_L), n$).\\
$\Lambda \gets \mbox{Abstain}(\Lambda)$ \cite{varma2018snuba}\\
Sort $A$ in increasing order.\\
\For{$i = 1, ..., \|A\|$}{
Randomly select $A_i$ points $X_P$ from $X_U$.\\
$X'_L \gets X_L \cup X_P$\\
$Y'_L \gets Y_L \cup \Lambda(X_P)$\\
Train downstream classifier $C_i$ with $(X'_L, Y'_L)$.\\
}
\caption{{\algname} for Weak Supervision.} \label{alg:weak_sup}
\end{algorithm}

%% file: sec_experiments.tex
\section{Experiments}~\label{sec:experiments}
We evaluate {\algname} in multiple real world behavior analysis domains (Section~\ref{sec:datasets}), and show that our framework outperforms existing LF generation methods in weak supervision and active learning settings (Section~\ref{sec:main_results}). 
Since researchers often study multiple behaviors in a domain~\cite{segalin2020mouse,eyjolfsdottir2014detecting}, we consider each behavior its own task.

\subsection{Datasets}~\label{sec:datasets}
We use datasets from behavioral neuroscience (mouse and fly behaviors) as well as sports analytics (basketball player trajectories).
These datasets include rare behaviors, multi-behavior tasks, and sequential data, making them good representations of real-world behavior analysis tasks.
Each dataset contains a train, validation, and test split; the validation split is only used for model checkpoint selection.

\textbf{Fly vs. Fly} (Fly).
The fly dataset~\cite{eyjolfsdottir2014detecting} contains frame-level annotations of videos of interactions between two fruit flies. 
Our train, validation, and test sets contain 552k, 20k, and 166k frames. 
We use fly trajectories tracked by FlyTracker~\cite{eyjolfsdottir2014detecting} and evaluate on 6 behaviors: lunge, wing threat, tussle, wing extension, circle, copulation.
This is a multi-label dataset and we report the mean Average Precision (mAP) over binary classification tasks for each behavior. 
All behaviors except for copulation are rare; lunge, wing threat, and tussle occur in $<5\%$ of frames, and wing extension and circle occur in $<1\%$ of frames. 
The domain-level LFs for this dataset are based on features from~\cite{eyjolfsdottir2014detecting}.

\textbf{CalMS21} (Mouse). The CalMS21 dataset~\cite{sun2021multi} consists of frame-level pose and behavior annotations from videos of interactions between pairs of mice. 
We use data from Task 1 (532k train, 20k validation, 119k test) and evaluate on a set of 3 behaviors: attack, investigation, and mount.
These behaviors are mutually exclusive and we report the mAP over these classes. 
We use a subset of the features in ~\cite{segalin2020mouse} as domain-level LFs for this dataset.

\textbf{Basketball}.
The Basketball dataset, also used in \cite{yue2014learning, shah2020learning, zhan2018generating}, contains sequences of basketball player trajectories from Stats Perform (18k train, 1k validation, 2.7k test). 
Labels for which offense player (5 total) had the ball for the majority of the sequence were extracted with ~\cite{ballhandler}.
We perform sequential classification in downstream tasks, and report the mAP over each offense player vs. the other 4.
Our domain-level LFs include player acceleration, velocity, and position among others. 
We exclude information about the ball position in the domain-level LFs and data features to focus on analyzing player behaviors.

\subsection{Baselines}
We compare {\algname} to two main baselines: student networks from student-teacher training and decision trees from SNUBA~\cite{varma2018snuba}.
We show that {\algname} outperforms both in data efficiency, requiring a fraction of the data to achieve or exceed performance parity.
For both baselines, domain-level LFs are incorporated as input features to evaluate the effectiveness of {\algname} and not the domain-level LFs themselves.
We do not compare against IWS \cite{boecking2020interactive}, as IWS is a human-in-the-loop LF generation system.
We also do not compare against ASTRA \cite{karamanolakis2021self}, as ASTRA is a weak supervision framework for using \textit{task-level} LFs in self training. 
However, ASTRA can be used as a downstream task for {\algname}.

\textbf{Student Networks}
Student-teacher training (from knowledge distillation\cite{noisystudent}) has been used successfully in self-training.
We adopt the concept of student networks by training models with similar capacity as the downstream classifier to serve as LFs.
In weak supervision experiments, these student LFs and the label model (Equation \ref{eqn:label_model}) serve as a teacher model for the downstream classifier.

\textbf{Decision Trees and SNUBA}
Decision trees have been shown to be good LFs~\cite{varma2018snuba} and offer some degree of interpretability.
The SNUBA framework~\cite{varma2018snuba} generates a diverse set of decision tree LFs by training $2^{k} - 1$ decision trees over all feature subsets and then pruning trees based on a diversity and performance metric, where $k$ is the feature dimension of $\mathcal X$.
Clearly, this is intractable for large $k$, which is often the case for behavior analysis tasks.
Furthermore, SNUBA does not use domain knowledge, instead relying on the complete set of decision trees for data efficiency.
In relation to SNUBA, {\algname} can be viewed as an scalable alternative to the synthesizer and pruner stages.

\subsection{Training Setup}
Our experimental setup consists of two stages: obtaining LFs, and evaluating generated LFs in downstream tasks.
Our downstream tasks include active learning, where LFs are used to select data for labeling, and weak supervision, where LFs generate pseudolabels for unlabeled data points.

\subsubsection{Obtaining labeling functions}
\textbf{Synthesized Programs via {\algname}}. 
For each domain, we use a simple DSL that includes add, multiply, fold, and differentiable if-then-else (ITE) structures among others.
We synthesize programs with our diverse program synthesizer and {\astar} search.
Our cost function is the sum of the $F_1$ cost from~\cite{shah2020learning} and our diversity cost $C_{P, \mathcal P}$.
We set $q(x)$ to $x^2$ and $m$ to $\log_2{\|\Lambda_m\|}$.
Program parameters are trained with weighted cross entropy loss.
More information about the exact DSL used is in the Supplementary Materials.

\textbf{Student Networks}.
We use neural networks for frame classification tasks and LSTMs for scene classification tasks.
To induce diversity in the learned student networks, we take inspiration from \cite{noisystudent} and randomly set the size of each layer so the ``expected'' student network is of similar capacity as the downstream classifier.
All student networks are trained using weighted cross entropy loss.

\textbf{Decision Trees}.
We fit decision trees using Gini impurity as the split criteria. 
We limit the depth of decision trees to $\log_2{k}$, so the number of nodes is $O(k)$.
We select diverse sets of decision trees by pruning a superset of trees based on coverage and performance, similar to how SNUBA does\cite{varma2018snuba}. 
However, unlike SNUBA, we group our features when generating the superset, as training $2^k - 1$ decision trees is intractable with our datasets.

\subsubsection{Downstream Tasks}
We use 3 LFs in our main experiments. Experiments with more LFs (5, 7) are in the Supplementary Materials.

\textbf{Active Learning}.
As previous described, we evaluate the performance of {\algname} at multiple data amounts, selecting additional labeled data with active learning at each amount (Algorithm \ref{alg:active_learning}).
We use max-entropy uncertainty sampling on downstream classifier outputs to select points for labeling~\cite{lewis1996uncertainty}.
We use $\{$1000, 2000, 3500, 5000, 7500, 12500, 25000, 50000$\}$ frames for the fly and mouse datasets and $\{$500, 1000, 1500, 2000, 3000, 4000, 5000$\}$ sequences for the basketball dataset.

\textbf{Weak Supervision}.
In our weak supervision experiments, we use factor graph model proposed in \cite{ratner2016data, ratner_metal}.
\begin{equation}
p_\theta(Y_U, \Lambda) = Z_\theta^{-1} \exp\Big( \sum_{i = 1}^{\|X_U\|} \theta^T \phi_i(\Lambda(X_{U_i}), Y_{U_i})\Big).
\label{eqn:label_model}
\end{equation}
Here, LF accuracies are modeled by factor $\phi_{i, j}^{Acc}(\Lambda, Y_U) = \mathbb{1}\{\Lambda_j(X_{U_i}) = Y_{U_i}\}$, and the proportion of data the LF labels is modeled by $\phi_{i, j}^{Lab}(\Lambda, Y_U) = \mathbb{1}\{\Lambda_j(X_{U_i}) \neq \emptyset\}$.

For the labeled dataset, we use 2000 frames for the fly and mouse datasets, and 500 sequences for the basketball dataset. 
Our unlabeled data amounts are set to $\{1\times, 2\times, 3\times, 4\times, 5\times\}$ the number of labeled points.

\begin{figure*}[t]
\centering
\includegraphics[width=0.33\linewidth, clip]{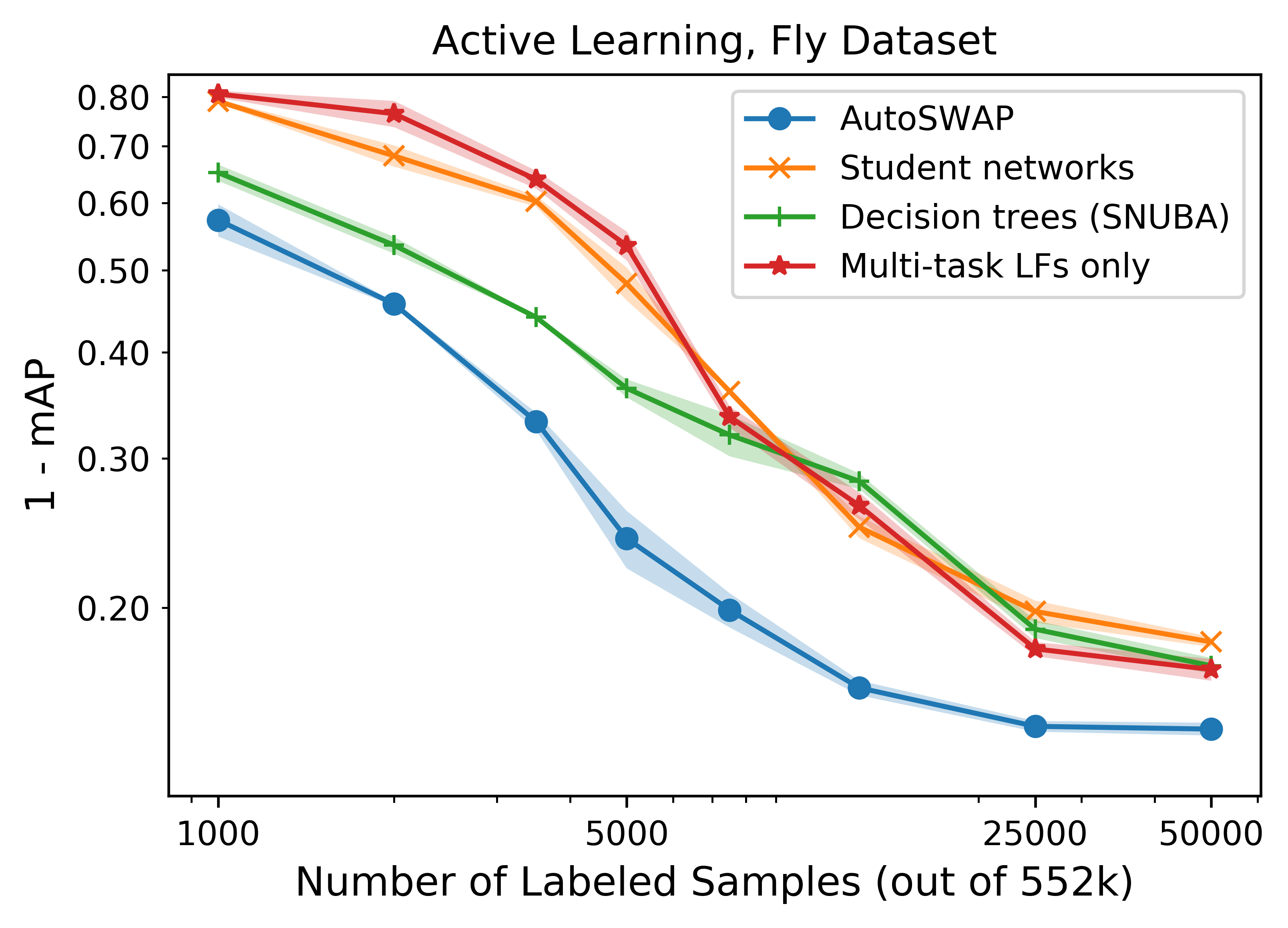}
\includegraphics[width=0.33\linewidth, clip]{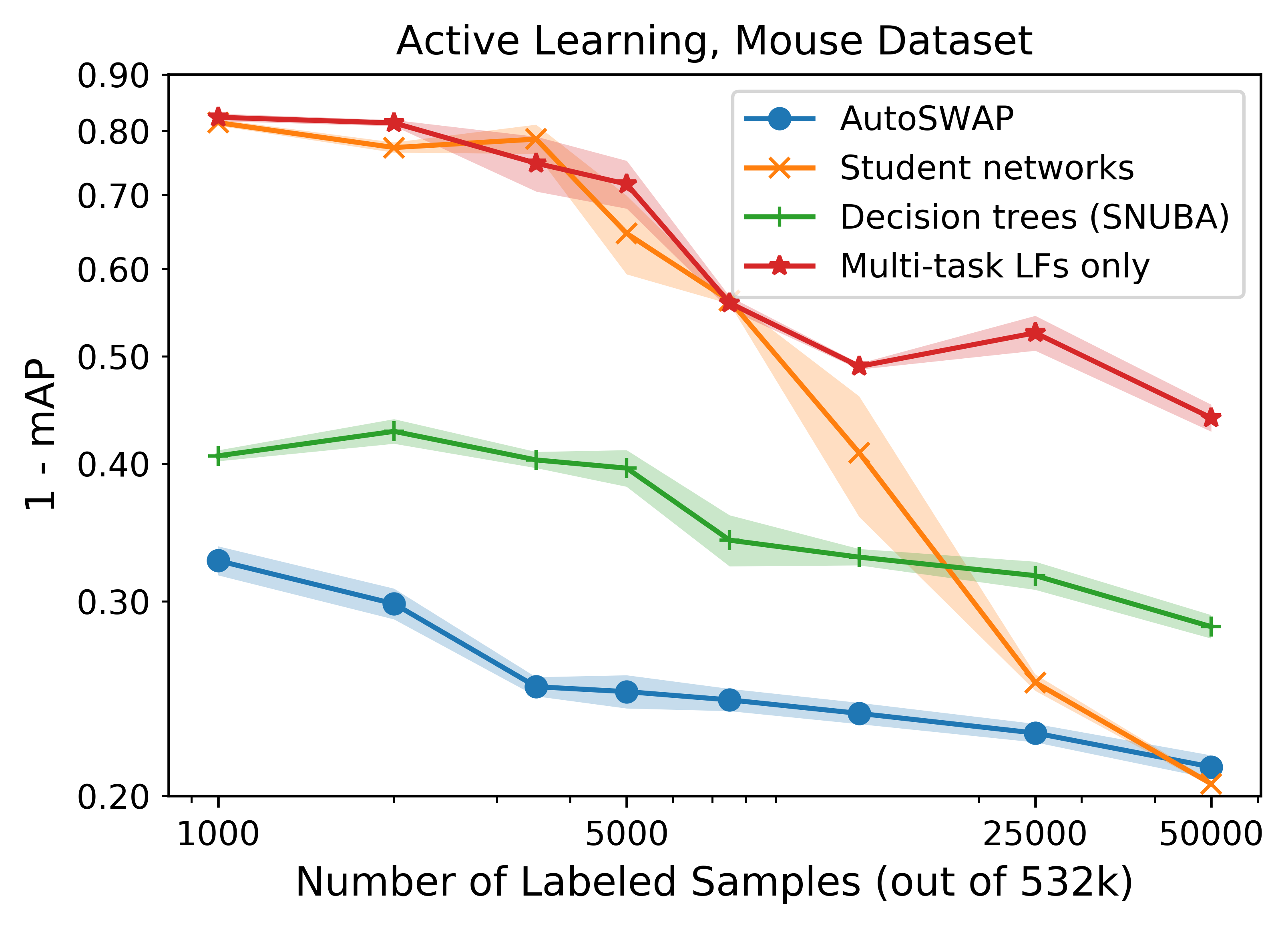}
\includegraphics[width=0.33\linewidth, clip]{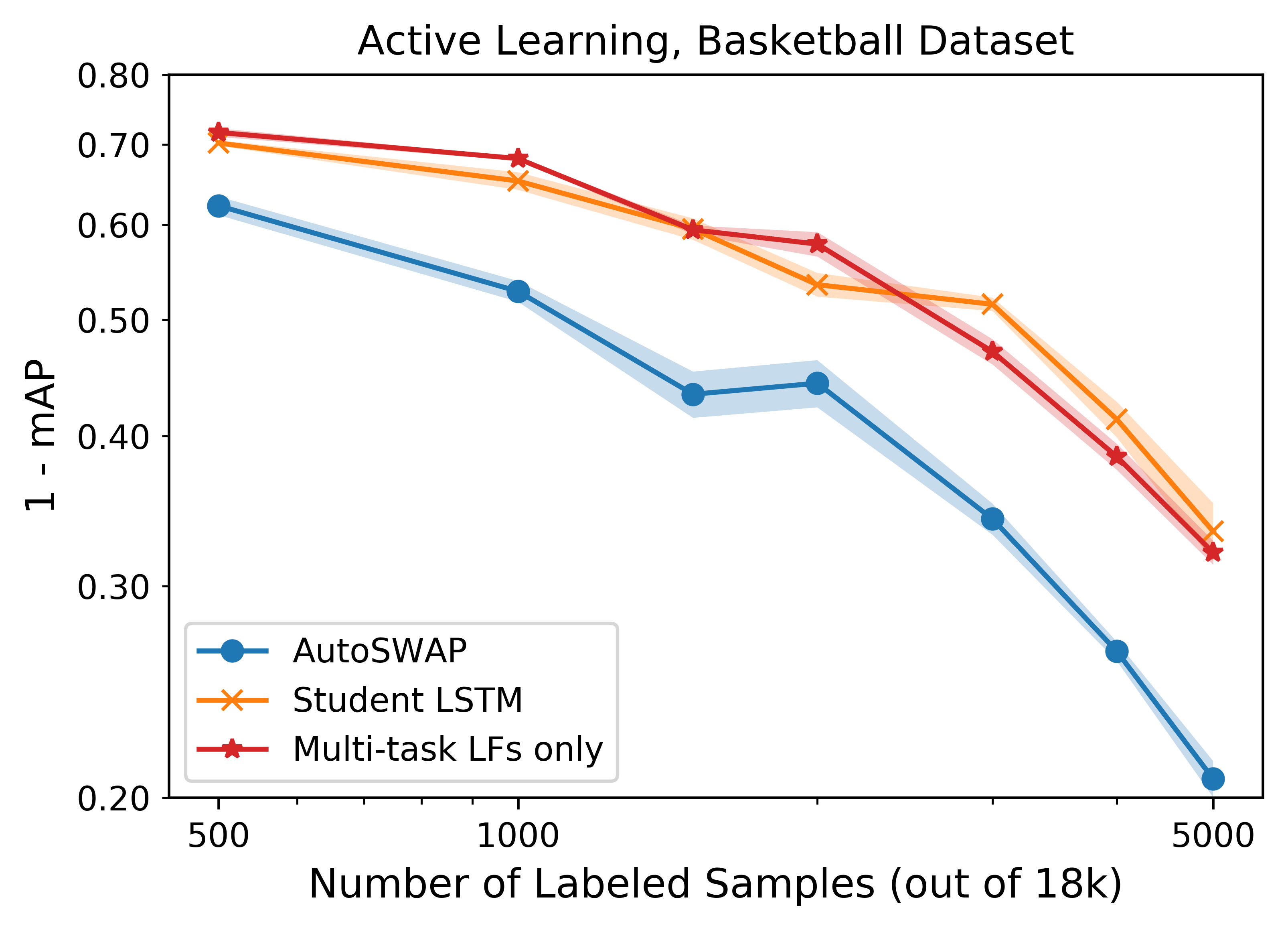}\\
\includegraphics[width=0.33\linewidth, clip]{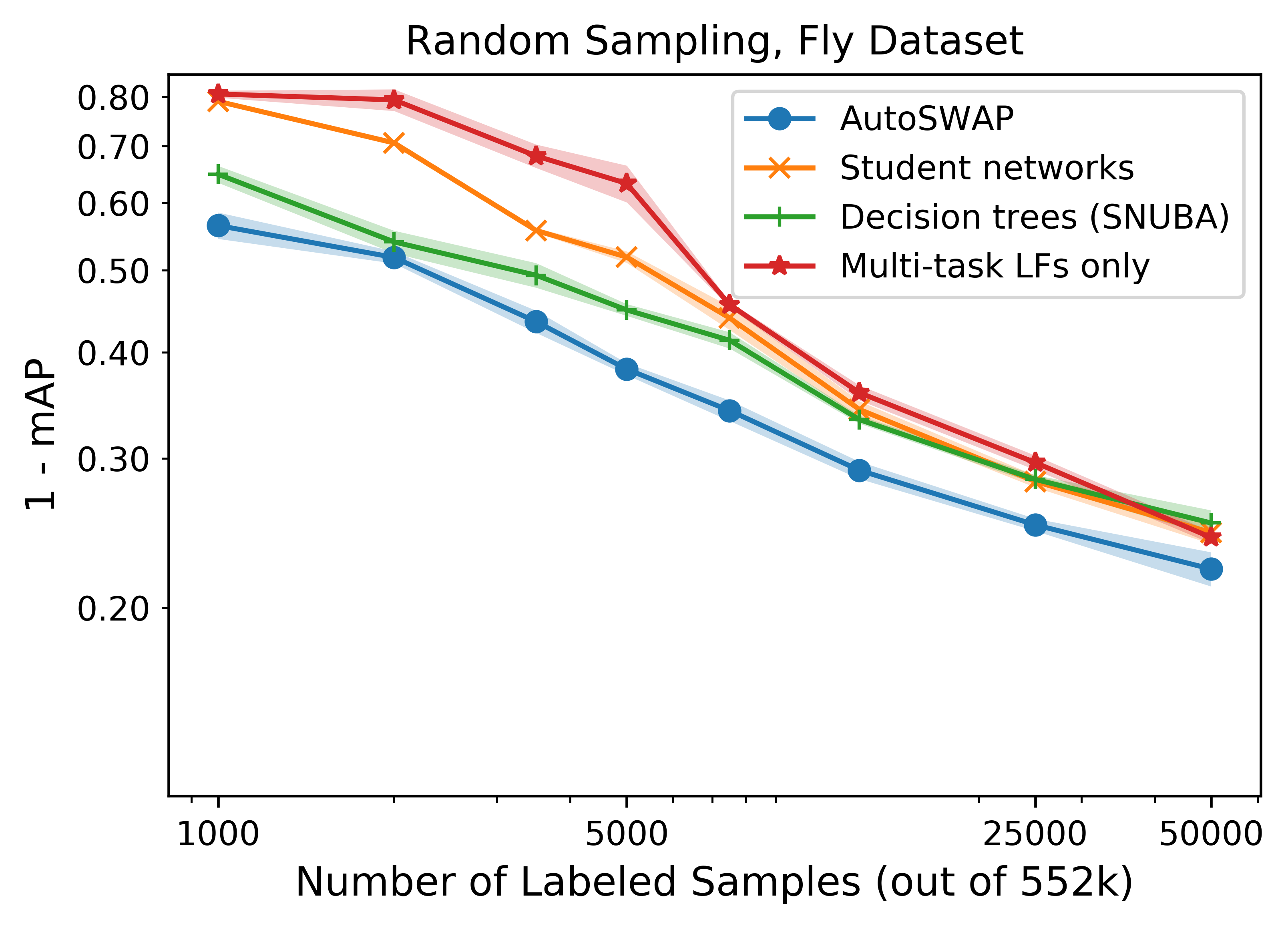}
\includegraphics[width=0.33\linewidth, clip]{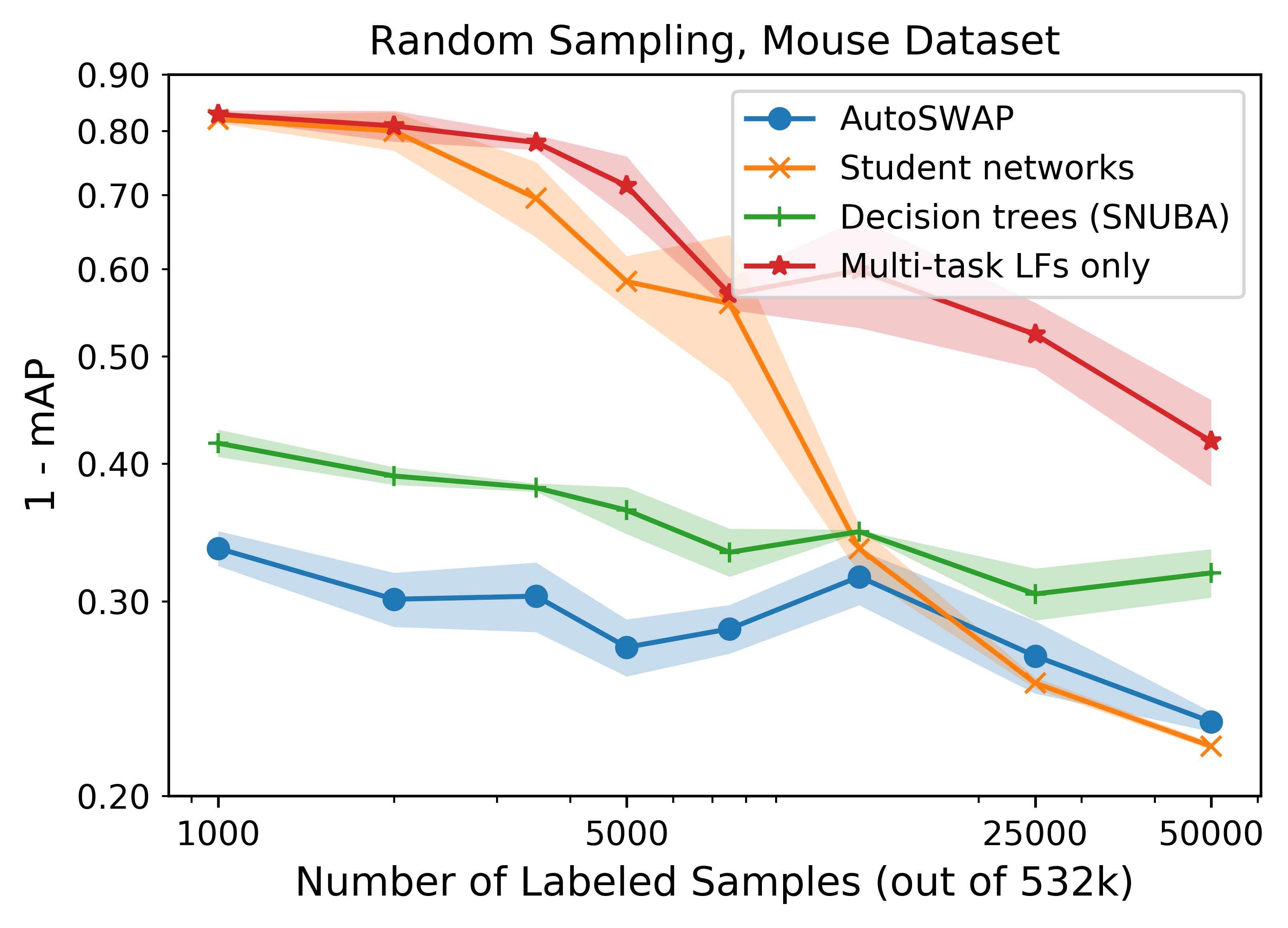}
\includegraphics[width=0.33\linewidth, clip]{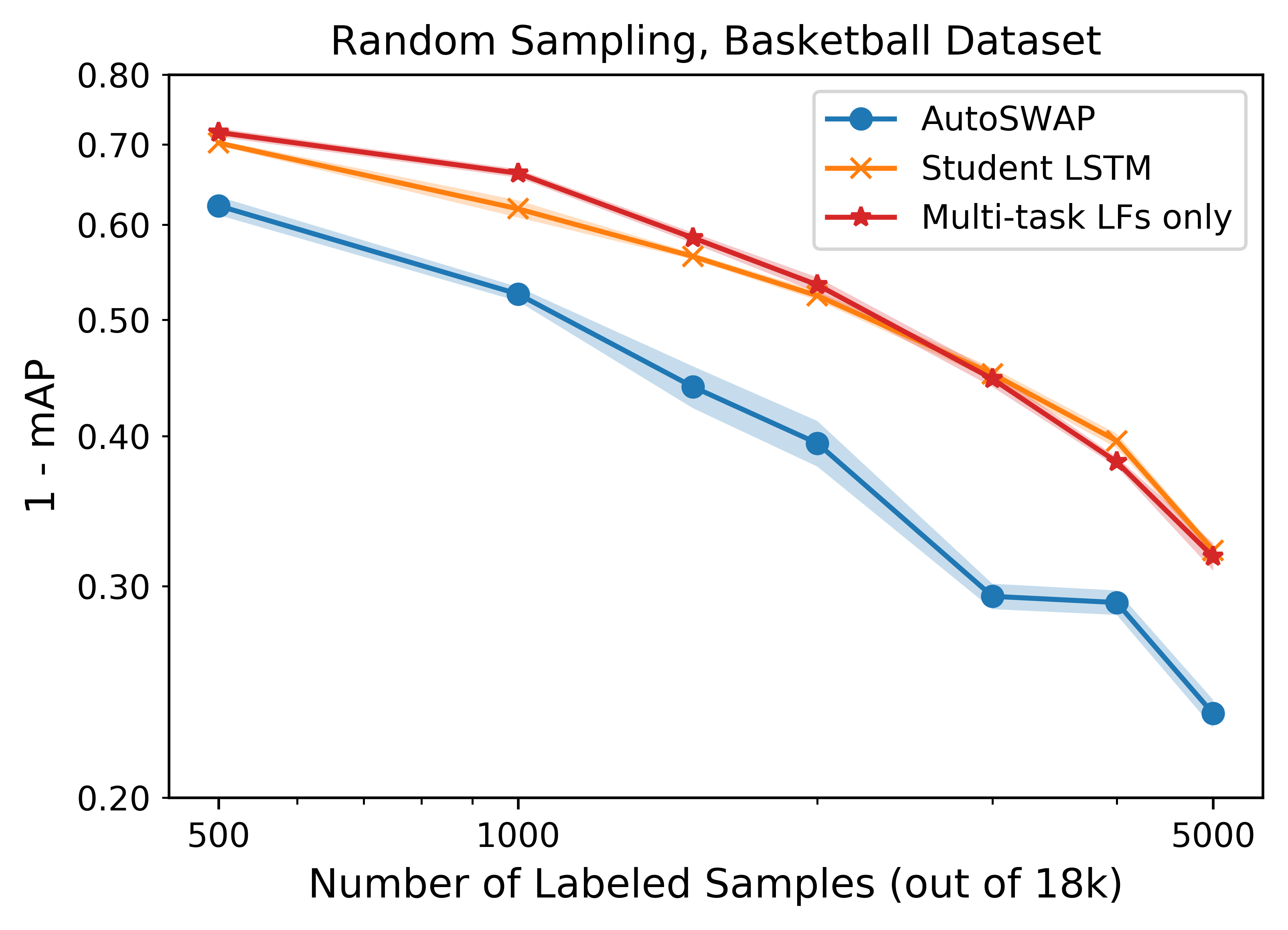}
\caption{{\algname} Active Learning Experiments. Each line represents the mean of 5 random seeds for an automatic labeling function method. The shaded region is the standard error of the seeds. As can be seen, {\algname} matches or outperforms all baseline methods using only a fraction of the data. Note that all plots are on log-log scales.} \label{fig:active_learning}
\end{figure*}

\begin{figure*}[t]
\centering
\includegraphics[width=0.33\linewidth, clip]{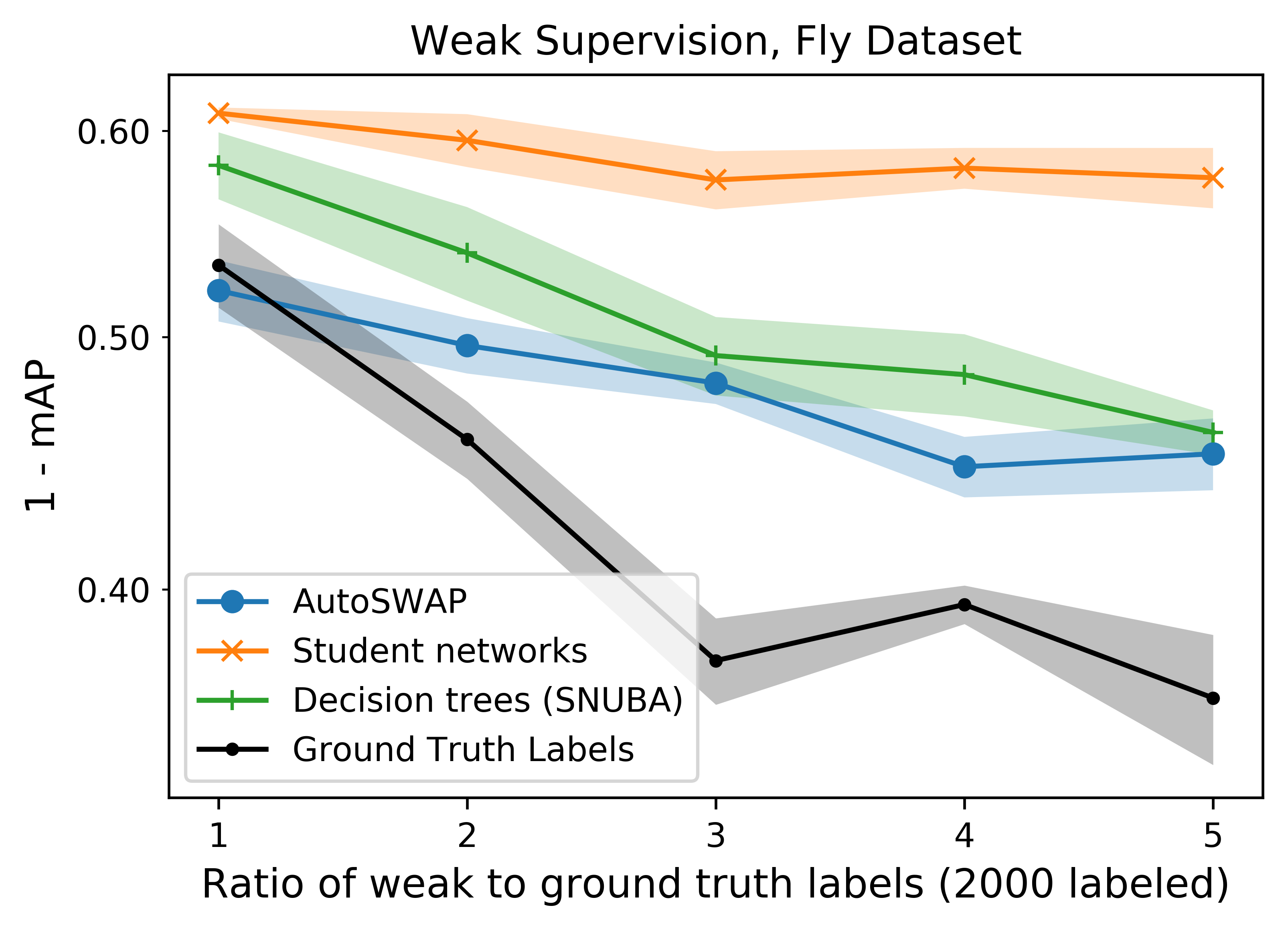}
\includegraphics[width=0.33\linewidth, clip]{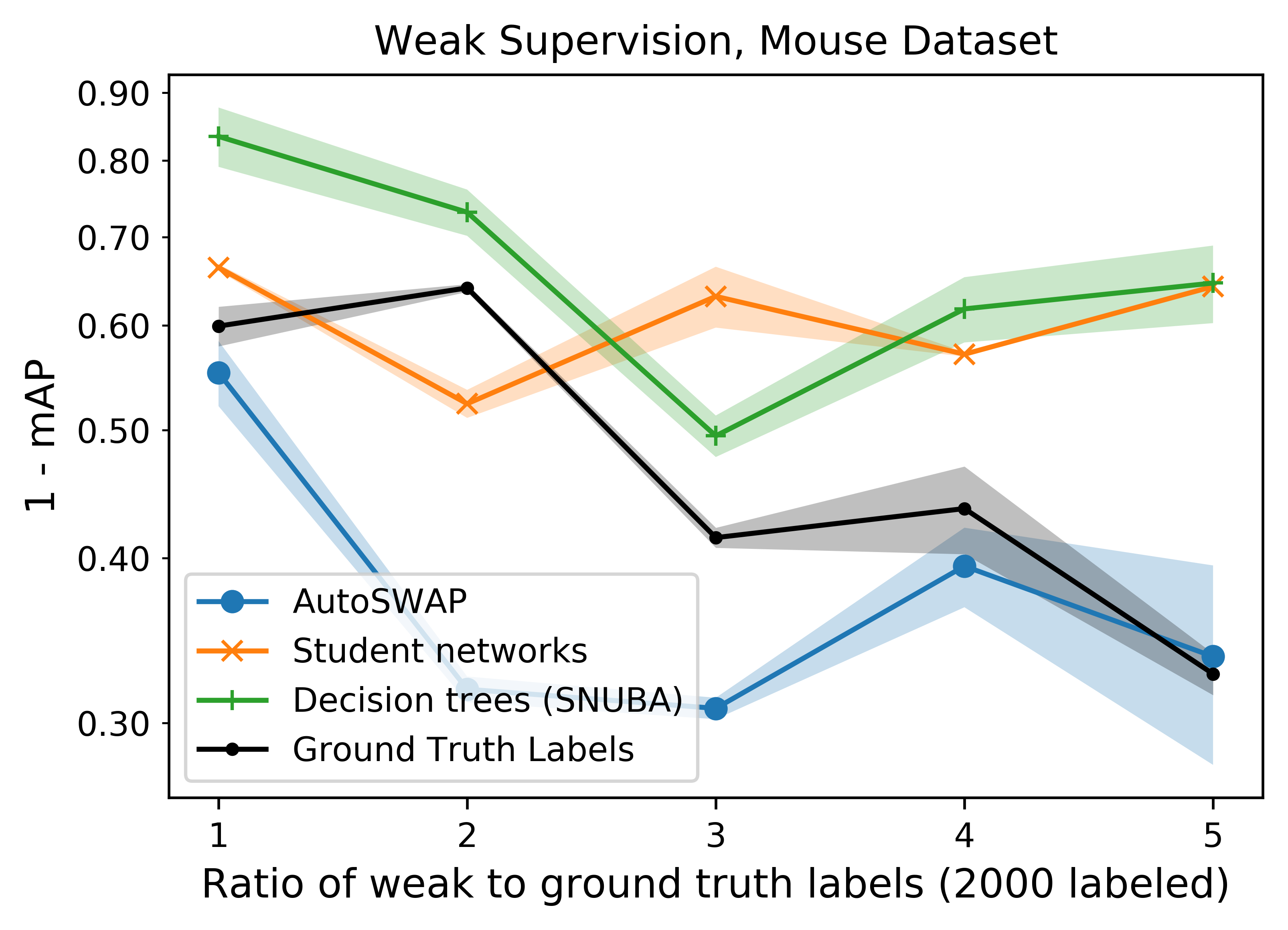}
\includegraphics[width=0.33\linewidth, clip]{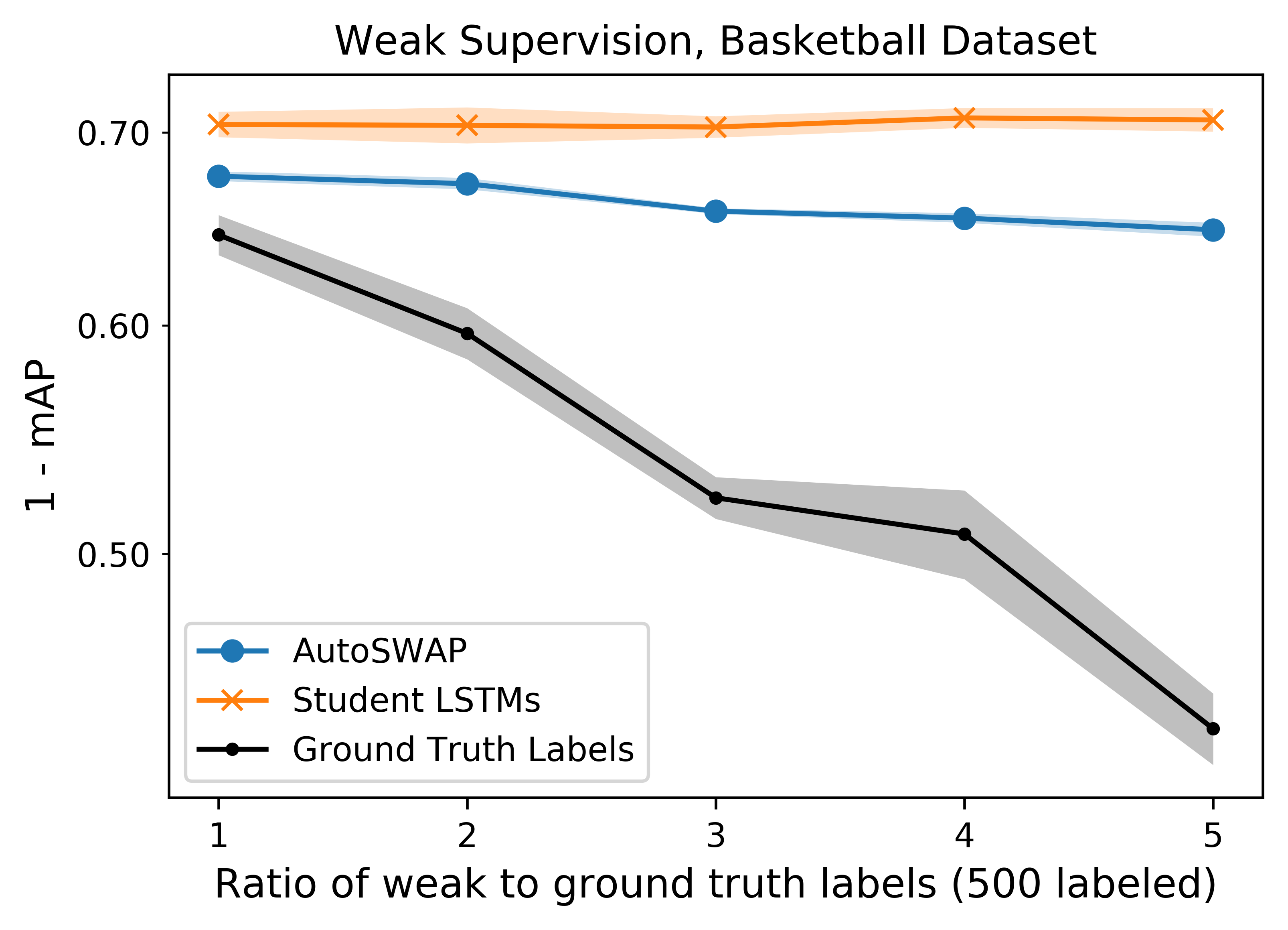}
\caption{{\algname} Weak Supervision Experiments. Each line represents the mean of 5 random seeds for an automatic labeling function method. The shaded region is the standard error of the seeds. The gray line shows performance when ground truth labels are used as weak labels. Although it may seem odd that {\algname} outperforms ground truth labels in the Mouse dataset, weak labels have been observed to outperform ground truth labels in other works \cite{karamanolakis2021self}. Note that all plots are on log-log scales.}
\label{fig:weak_supervision}
\end{figure*}

\section{Results}
We compare the data efficiency of {\algname} against the baselines on our behavior analysis datasets. We do not run the decision tree (SNUBA) baseline on the Basketball dataset as it contains only sequential data.

\subsection{Data Efficiency Results}~\label{sec:main_results}
\textbf{Active Learning}.
{\algname} LFs are far more data efficient than baseline methods across all datasets, indicating that {\algname} is effective in reducing label cost in active learning settings (Figure \ref{fig:active_learning}). 
This difference is especially pronounced in the Mouse dataset, where {\algname} achieves parity with decision tree LFs with roughly 30$\times$ less data.
In the Fly dataset, {\algname} is consistently $\sim4\times$ more data efficient than the baselines, and no baseline is able to reach performance parity with {\algname} by 50000 samples (9.1\% of the entire Fly dataset).
We observe a similar trend in the Basketball dataset, with {\algname} being $\sim2\times$ as data efficient.
We also observe an improvement in data efficiency even when using random sampling, and note that uncertainty sampling widens the gap between {\algname} and the baselines.

While {\algname} LFs themselves do not necessarily perform better than baseline LFs when evaluated on their own (see the Supplementary Materials), they do provide a stronger learning signal for downstream classifiers than the baselines.
These data efficiency differences can be attributed in part the structural domain knowledge encoded in the DSL, as the domain-level LFs themselves perform significantly worse.
For example, a {\algname} LF classifying ``lunge vs. no behavior'' for the Fly dataset can be seen in Figure \ref{fig:prog2tree}, and the structure of this program cannot be easily approximated with a decision tree or a neural network.

\textbf{Weak Supervision}.
Similar to our active learning experiments, we observe that AutoSWAP is more data efficient than the baselines in weak supervision settings (Figure \ref{fig:weak_supervision}).
We note that the ground truth labels are not a baseline in this setting, as they are essentially an ``optimal'' case where the weak labels match the ground truth labels.

On the Fly dataset, AutoSWAP generally performs better than both baselines, and on the Mouse and Basketball datasets, no baseline is able to match the performance of {\algname} LFs at any evaluated amount of annotated data.
{\algname} is even able to outperform the ground truth labels in the Mouse dataset at some levels of annotated data, which indicates that the learned LFs are especially informative.
Finally, we observe that {\algname} generally improves with more weakly labeled data points, which is useful as there is no expert annotation cost to using more weakly labeled data points.

\subsection{Additional Results}
 
\begin{figure}
\centering
\includegraphics[width=0.9\linewidth]{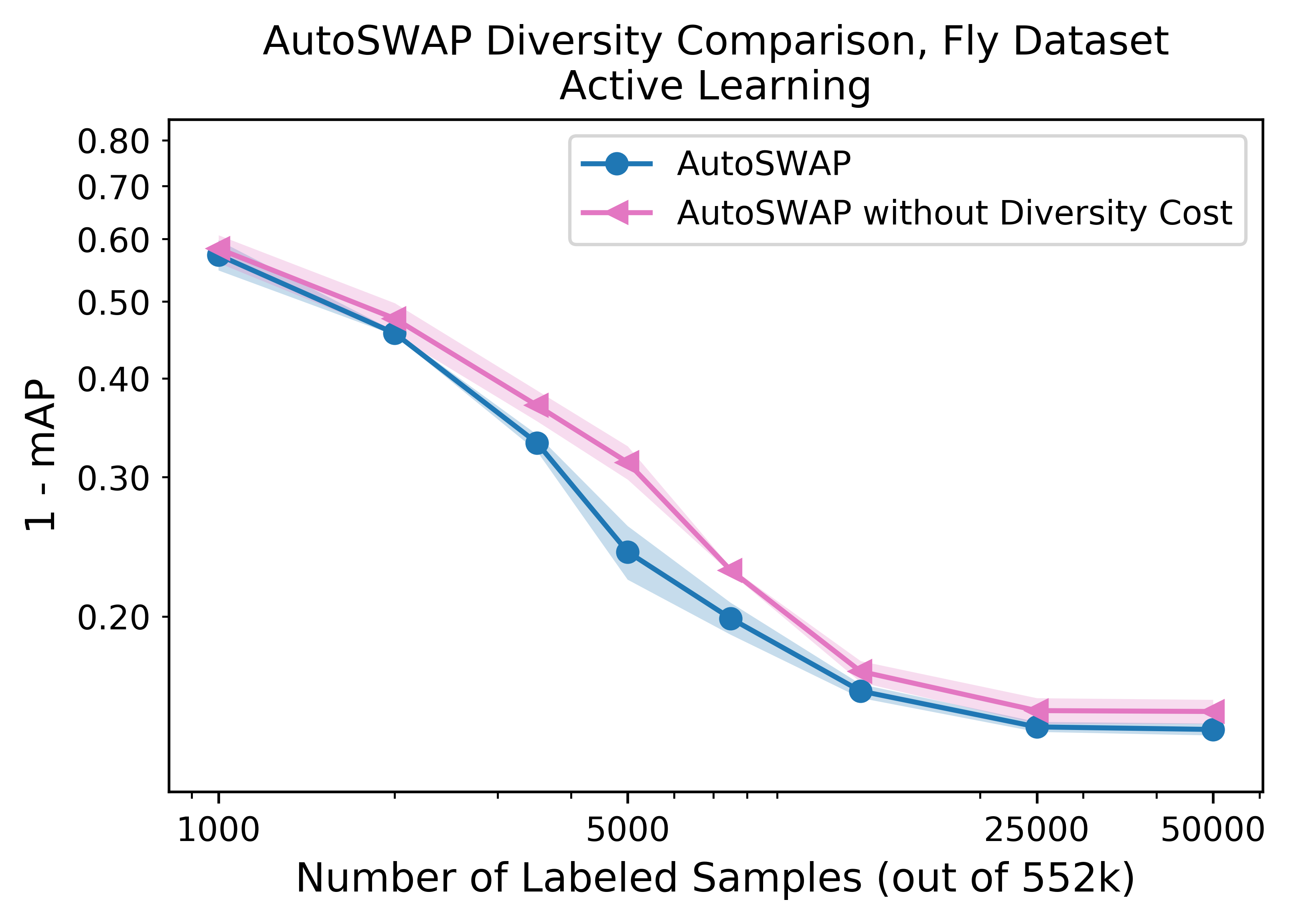}
\includegraphics[width=0.9\linewidth]{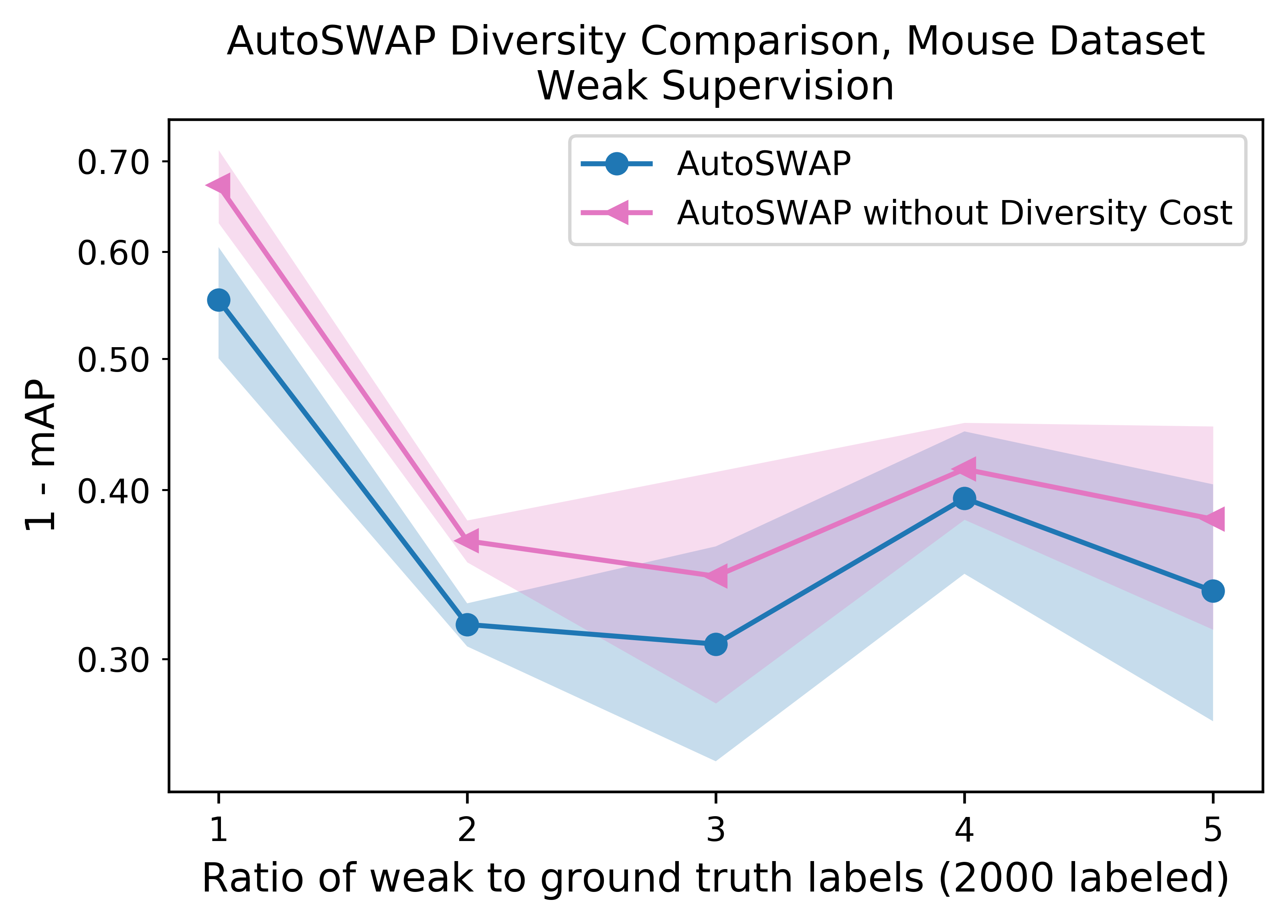}
\vspace{-0.1in}
\caption{Diversity Cost Utility Comparison. Synthesizing diverse sets of programs instead of purely $\epsilon$-optimal sets improves AutoSWAP, showing the utility of the structural diversity cost.}
\label{fig:div_comp}
\end{figure}

\textbf{{\algname} Diversity Cost}. 
The diversity cost is an important part of {\algname}.
As can be seen in Figure \ref{fig:div_comp}, synthesizing purely optimal programs w.r.t. Equation \ref{eqn:optimal_diff_prog} results in worse performance than synthesizing diverse sets of programs.
This mirrors the observations in \cite{varma2018snuba}, where using diverse sets of decision trees improves performance.

\textbf{Interpretability of Labeling Functions}.
An important part of behavior analysis is being able to interpret learned models. 
Neural networks and LSTMs are by nature not interpretable. 
Decision trees offer some degree of interpretability, but are limited to branched if-then-else statements.
With {\algname}, complex yet interpretable programs can be learned by using interpretable structures in the DSL (Figures \ref{fig:prog2tree}, \ref{fig:example_task_lfs}, Supplementary Materials).

\textbf{Effect on Rare Behaviors}. 
Rare behaviors can be difficult to analyze, as even with large datasets very little data exists.
Our fly domain results show that {\algname} greatly improves data efficiency for rare behaviors, as 5 of the 6 behaviors we study occur in $<5\%$ of the frames.
We note the copulation task (which is not rare) does not bias our Fly domain data efficiency comparison as all tested methods achieve near-perfect performance on it.

\begin{figure}
\centering
\lstset{
  basicstyle=\footnotesize,
  frame=tb,
  numbers=none,
  backgroundcolor=\color{light-gray}
}
\begin{lstlisting}
Fly Domain:
    (Lunge) Map(Add(Multiply(Speed, WingRatio), Positional))
    (Tussle) Map(SimpleITE(Angular, Speed, WingDistance))))
Mouse Domain:
    Map(Fold(SimpleITE(DistanceM1, AngleM1, SpeedM1)))
    Map(SimpleITE(PositionalM2, SpeedM1, DistanceM1))
Basketball Domain:
    Fold(Add(PlAccel(), Add(PlPos(), PlVel())))
    Fold(SimpleITE(BVel(), PlVel(), PlPos()))
\end{lstlisting}
\vspace{-0.1in}
\caption{Example {\algname} task-level LFs (architectures only). LFs are composed of domain-level LFs and structural relations from the DSL. For example, the ``Fly Lunge LF'' labels whether a fly is lunging using the fly's speed, wing ratio, and positional domain-level LFs. More detailed descriptions of {\algname} LFs can be found in the Supplementary Materials.}
\label{fig:example_task_lfs}
\end{figure}

%% file: sec_conclusion.tex
\section{Discussion and Conclusion}

We propose {\algname}, a framework that uses program synthesis to automatically synthesize diverse LFs.
Our results demonstrate the effectiveness of our framework in both active learning and weak supervision settings and across three behavior analysis settings.
We find that with existing domain-level LFs \cite{eyjolfsdottir2014detecting,segalin2020mouse} and a simple DSL, {\algname} can synthesize highly data efficient task-level LFs with minimal amounts of labeled data, thus reducing annotation requirements for domain experts.

Additionally, we introduce a novel structural diversity cost and admissible heuristic for synthesized programs, which allows {\algname} to scalably synthesize diverse LFs with informed search algorithms.
This further improves the performance of our framework in behavior analysis settings, all without requiring domain experts to repeatedly hand-craft task-level LFs.
Overall, {\algname} effectively integrates weak supervision with behavior analysis, and greatly reduces domain expert effort through automatically synthesizing task-level LFs from domain-level knowledge.


\textbf{Limitations.}
While our DSL and LFs are at the domain-level, our method requires task-level information in the form of a small labeled dataset to synthesize LFs. Additionally, the LFs provided by domain experts should be informative of behavior (although we do show that current behavioral features~\cite{eyjolfsdottir2014detecting,segalin2020mouse} studied by domain experts are sufficient for this task). Extensions to automate other aspects of our framework while taking into account domain expert knowledge, such as library learning~\cite{NEURIPS2018_7aa685b3} or integrating perception~\cite{NEURIPS2018_edc27f13}, may further reduce expert effort. However, we note that our current framework already leads to significant reductions in data requirements.  


\textbf{Societal Impact.}
Automatically generating interpretable LFs to reduce expert effort can help behavior analysis across domains, such as in neuroscience, ethology, sports analytics, and autonomous vehicles, among others.
Our framework leverages inductive biases in the DSL to produce interpretable programs; however, since humans create the DSL, interpret programs, and annotate data, users should be aware of potential human-encoded biases in these steps. Additional care is especially needed in human behavior domains, such as with informed consent of participants and responsible handling of data.



%% file: supp_implementation.tex
\section{Implementation Details}~\label{sec:implementation}
\subsection{Datasets and Domain-level Labeling Functions}

Each of our behavior domains contains a dataset with tracked keypoints and features, domain-level labeling functions (LFs), and a DSL. Below, we provide more details about the datasets and domain-level LFs. Details about the DSL are provided in Section \ref{sec:dslsupp}.

\textbf{Fly vs. Fly}.
The Fly dataset~\cite{eyjolfsdottir2014detecting} consists of videos ($144 \times 144$ at 30 Hz), extracted behavioral features based on fly trajectories, and frame-level behavior annotations.
We use the ``Aggression" and ``Courtship" videos from this dataset, similar to~\cite{sun2020task}, and use the frame-level behavior annotations to evaluate our framework for behavior classification.
The breakdown of behaviors in this dataset is given by: lunge - 1.24\%, wing threat - 4.26\%, tussle - 0.35\%, wing extension - 3.31\%, circle - 0.80\%, copulation - 59.73\%, and no behavior - 30.42\%.
This dataset is available under the CC0 1.0 Universal license.

\begin{figure*}
\centering
\includegraphics[width=\textwidth]{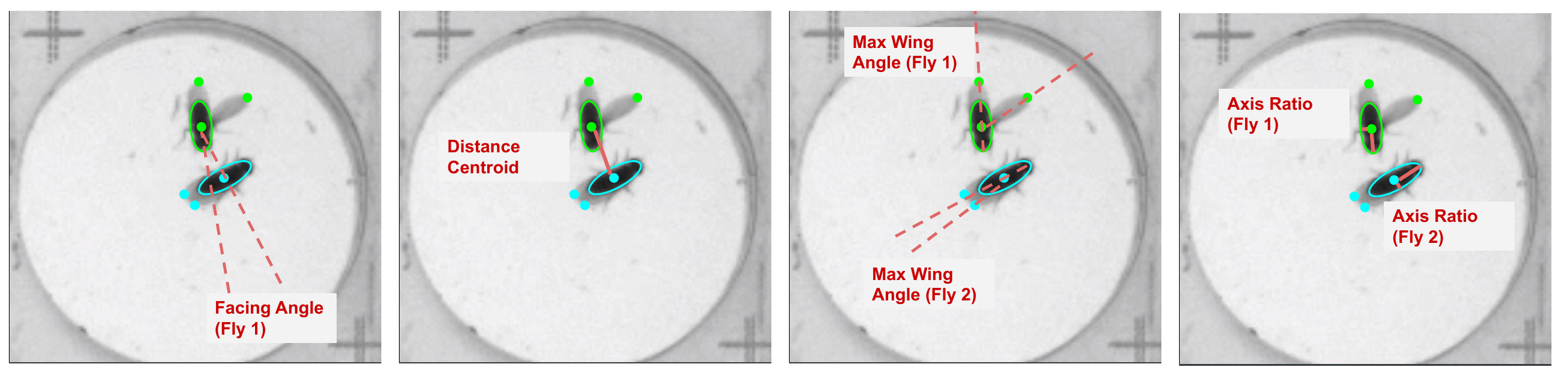}
\caption{Visualizing a subset of programs computed on  Fly vs. Fly. Reproduced with permission from~\cite{sun2020task}.}
\label{fig:fly_programs}
\end{figure*}

The domain-level LFs for the Fly DSL are based on the behavioral features provided with the dataset, and are crafted from the outputs of FlyTracker~\cite{eyjolfsdottir2014detecting}.
They are similar to the fly behavior features used in~\cite{sun2020task}, and we provide visualizations for example LFs in Figure \ref{fig:fly_programs}.
The domain-level LFs included in the Fly DSL, grouped by category, are:
\begin{itemize}
    \item Angular LFs: angular velocity of each fly, angle between fly trajectories, facing angle of flies.
    \item Linear (speed) LFs: linear velocity and speed of each fly.
    \item Positional LFs: position of each fly, distance between flies, distance between the legs of each fly.
    \item Ratio LFs: ratio of major and minor axis of fly shape for each fly, ratio of major and minor axis of fly body, ratio of major and minor axis of bounding box around wing spread.
    \item Wing LFs: minimum and maximum wing angle (window), mean wing length (window), wingspan.
\end{itemize}
where ``window'' indicates that the LF is computed over a sliding window of the past 20 frames.

\textbf{CalMS21 (Mouse)}. CalMS21~\cite{sun2021multi} consists of trajectory data, frame-level behavior annotations, and video data ($1024 \times 570$ at 30 Hz) of a pair of interacting mice in a resident-intruder assay. The trajectory data from CalMS21 is from the MARS detector~\cite{segalin2020mouse}, which detects 7 anatomically-defined body parts on each mouse in the form of keypoints.
We use the trajectory data and behavior annotations from the Task 1 train/test split to evaluate our framework for mouse behavior classification.
The breakdown of behaviors in this dataset is given by: attack - 3.44\%, investigation - 27.56\%, mount - 7.45\%, and other - 61.56\%
This dataset is available under the CC-BY-NC-SA license.

\begin{figure*}

\centering
\includegraphics[width=\textwidth]{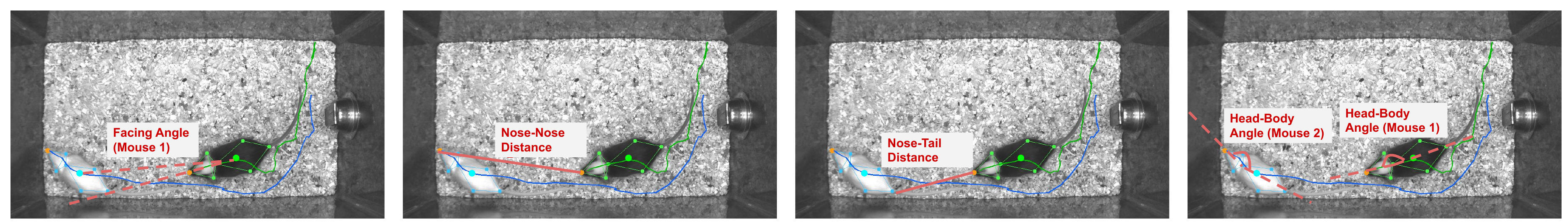}

\caption{Visualizing a subset of programs computed on  CalMS21. Reproduced with permission from~\cite{sun2020task}.}
\label{fig:mouse_programs}

\end{figure*}

The domain-level LFs for the Mouse DSL are based on existing behavioral features used by domain experts in \cite{segalin2020mouse} and are similar to the features used in~\cite{sun2020task} (visualized in Figure~\ref{fig:mouse_programs}). The LFs included in the Mouse DSL, grouped by category, are:
\begin{itemize}
    \item Positional LFs: nose, right ear, left ear, neck, RHS, LHS, and tail position for each mouse.
    \item Centroid LFs: centroid position of entire mouse, head, hips, and body for each mouse. 
    \item Angular LFs: orientation of entire mouse, head, body, and angle between head and body (L\&R) for each mouse, as well as angle and facing angle between the two mice.
    \item Shape LFs: ratio of major and minor axis of mouse body and bounding ellipse for each mouse, as well as ratio of bounding ellipse areas between the two mice, bounding ellipse overlap, and edge distance between bounding ellipses.
    \item Speed LFs: linear, radial, and tangential speed of each mouse, linear acceleration of each mouse, and relative speed of resident vs intruder mouse.
    \item Relative Distance LFs: relative distance between bodies, noses, heads, and centroids of the two mice, among other relative distances.
\end{itemize}

\textbf{Basketball}. 
The StatsPerform Generative Models Basketball \cite{yue2014learning} dataset consists of basketball player trajectories from NBA games. Each trajectory is located in the left half of the court, and contains 25 frames sampled at 3Hz of 5 defense players, 5 offense players, and 1 basketball.
As the Basketball dataset does not have ground truth labels for any tasks, we use a ``ballhandler'' label computed from the position of the ball relative to each player \cite{ballhandler}. As such, we exclude the position of the ball itself from our training data and domain-level LFs. The distribution of ballhandler by player is given by: 1 - 18.50\%, 2 - 22.09\%, 3 - 22.87 \%, 4 - 18.18\%, 5 - 18.36\%. Since we are interested in the ballhandler, we exclude the offense players from our training data and domain-level LFs.

The dataset itself is available for free on AWS Marketplace. The list of domain-level LFs included in the Basketball DSL, grouped by category, are:

\begin{itemize}
\item Player: acceleration, velocity, speed, and position of each defense player.
\item Ball: acceleration, velocity, and speed of the ball. Note this does not include the starting position of the ball so the true position of the ball cannot be computed from the velocity.
\end{itemize}

\subsection{Domain Specific Language (DSL)}
\label{sec:dslsupp}
The DSL we use in our experiments consists of elementary operations, operations on sequential data, and branching structures, among others. We use the same DSL for all domains; only the domain-level LFs differ from domain to domain. In Backus-Naur form, our DSL can be written as: 

\begin{equation}
\begin{array}{lll}
\alpha & ::= & x \mid c \mid +(\alpha_1,\dots, \alpha_k) \mid \cdot(\alpha_1, \dots, \alpha_k) \mid \times(\alpha_1, \dots, \alpha_k) \mid \frown(\alpha_1, \dots, \alpha_k) \\
&& \oplus_\theta (\alpha_1, \dots, \alpha_k) \mid \oplus_D~x \mid \ifc~\alpha_1~\thenc~\alpha_2~\elsec~\alpha_3 \\
&& \mapc~(\func~x_1, \alpha_1)~x \mid \foldc~(\func~x_1, \alpha_1)~c~x   
\end{array}
\end{equation}

Here, $x$ and $c$ represent input features and constants, respectively.
$+, \cdot, \times$, and $\frown$ represent the addition, dot product, outer product, and concatenation operators.
$\oplus_\theta$ represents domain-expert provided parameterized library functions and $\oplus_D$ represents domain-level labeling functions.
$\func~x. f$ represents a lambda that evaluates $f$ over $x$; for example, if $x$ is a sequence $\{x_0, x_1, \dots, x_k\}$ then $\func~x. f$ returns $\{f(x_0), f(x_1), \dots f(x_k)\}$.
$\mapc$ and $\foldc$ represent operations on vectors and sequences, respectively.

$\ifc-\thenc-\elsec$ represents a differentiable ITE construct, which can be written with our program notation $\Sem{\alpha}(x, \theta)$ as 
\begin{align}
y &= \sigma(\beta \Sem{\alpha}(x, \theta_1)) \\
\Sem{\ifc~\alpha_1~\thenc~\alpha_2~\elsec~\alpha_3}(x, (\theta_1, \theta_2, \theta_3)) &= y \Sem{\alpha_2}(x, \theta_2) + (1-y) \Sem{\alpha_3}(x, \theta_3) \label{eqn:diffite}
\end{align}
where $\sigma$ is the sigmoid function and $\beta$ serves as a temperature parameter. As $\beta \to \infty$, Equation \ref{eqn:diffite} becomes a regular ITE.

\subsection{Training Details}

\textbf{Student Networks}. 
We use neural networks (NNs) for nonsequential data and LSTMs for sequential data.
For NN based student networks, we use 3 hidden layers with layer sizes of $l_1 \sim \mathcal U\{50, 90\}, l_2 \sim \mathcal U\{50, 80\}, l_3 \sim \mathcal U\{20, 30\}$.
Each hidden layer is followed by a dropout layer with dropout probability $d_1 \sim \mathcal U_{[0, 0.2]}, d_2 \sim \mathcal U_{[0, 0.2]}, d_3 \sim \mathcal U_{[0, 0.1]}$.
All hidden layers use the ReLU activation function. 
For LSTM based student networks, we use a 1 or 2 layer LSTM with equal probability and $h \sim \mathcal U\{64, 128\}$ hidden units. If we use a 2 layer LSTM, we also set the LSTM dropout probability to $d \sim \mathcal U_{[0, 0.2]}$. The final hidden state of the LSTM is classified with a neural network with two hidden layers of sizes $l_1 \sim \mathcal U\{50, 80\}, l_2 \sim \mathcal U\{20, 30\}$. Both layers are each followed by a dropout layer $\sim_{i.i.d} \mathcal U_{[0, 0.2]}$ and use ReLU activation.
All student networks are trained with the Adam optimizer, using a learning rate of $10^{-4}$ for NNs and $3\times10^{-4}$ for LSTMs.

\textbf{AutoSWAP Programs}.
For the Fly and Mouse dataset, neural completions were trained for 6 epochs, and symbolic components were trained for 15 epochs. 
For the Basketball dataset, neural completions were trained for 4 epochs, and symbolic components were trained for 6 epochs. 
For all datasets, parameters were optimized with the Adam optimizer. A learning rate of $10^{-3}$ was used for the Fly and Mouse datasets, and $0.02$ was used for the Basketball dataset.
$\beta$ was set to 1 for the differentiable ITE construct.

\textbf{Downstream Classifier}.
Again, we use NNs for nonsequential data and LSTMs for sequential data. The downstream classifier NN is a 3 layer network with $(128, 64, 32)$ units in each hidden layer, respectively. Each layer is followed by a dropout layer with dropout probability $(0.5, 0.4, 0.3)$, respectively, and uses ReLU activation. For active learning experiments, the weak label outputs are also included via skip connections to each layer.
For LSTM based downstream classifiers, we use a two layer LSTM with 128 hidden units and a dropout probability of 0.2. The final hidden state of the LSTM is classified by a two layer neural network, with the first layer having 128 units with dropout probability 0.3, and the second layer having 48 units with dropout probability 0.2. Again, both layers use ReLU and weak label outputs are included via skip connections for active learning experiments.
All downstream classifiers are trained with the Adam optimizer, using a learning rate of $10^{-4}$ for NNs and $3\times10^{-4}$ for LSTMs.

\subsection{Reproducibility}

The code for our experiments is available at \href{https://github.com/autoswap/autoswap_cvpr_2022}{https://github.com/autoswap/autoswap\_cvpr\_2022}. All experiments were run on two different virtual machines with 12 Broadwell cores clocked at 2.6GHz and two Nvidia Tesla P40 GPUs each. Some experiment sets were run across different machines due to resource constraints, thus some results that should otherwise be identical have slight differences.

%% file: supp_programs.tex
\section{Learned Labeling Functions}~\label{sec:programs}
Here, we present some example AutoSWAP LFs and our interpretations of them. 
While domain experts may have different interpretations from us, the purpose of this section is to demonstrate the relative interpretability of AutoSWAP LFs.

\textbf{Example 1}: AutoSWAP labeling function for the Ballhandler task. The program can be interpreted as using the sum of the frame level probabilities ($\foldc$), with each frame level probability being determined from player velocities or player coordinates depending on the velocity of the ball. Summing over frame-level probabilities corresponds well with the ``majority'' part of the ballhandler task (find the player that was the ballhandler for the majority of the sequence). The ITE construct can be thought of as detecting passes, as the ball moves the fastest when it is being passed between players. Each $\textit{*Affine}$ labeling function contains a set of parameters describing a linear transformation on the base LF. We include the parameters for scalar affine LFs as $x_{[a, b]}$, which indicates the transformation $ax + b$ is applied. We do not include the parameters for vector LFs, as they are too large to list here, but examples can be found by running the code.
\begin{align*}
\foldc(&\func~x_t~.~[\\
&\ifc~\textit{BallVelocityAffine}(x_t)_{[0.73, -0.26]}\\
&\thenc~\textit{PlayerVelocitiesAffine}(x_t)\\
&\elsec~\textit{PlayerCoordinatesAffine}(x_t)\\
])~x_t&
\end{align*}

\textbf{Example 2}: AutoSWAP labeling function for the Wing Threat task (fly domain). The program can be interpreted as using the product of the resident fly's wing ratio and the sum of the resident fly's body ratio and the distance between the two flies to derive a probability for the resident fly displaying a wing threat towards the intruder fly. The program learns a positive $a$ for the wing ratio domain-level LF, which is reasonable as wing threats are correlated with wing spreading.
\begin{align*}
\mapc(&\func~x~.~[\\
&\times(\textit{WingRatioAffine}(x)_{[0.96, -0.71]}, \\
&\hspace{1cm}+(\textit{BodyRatioAffine}(x)_{[0.35, -0.20]}, \textit{FlyDistanceAffine}(x)_{[0.67, 0.13]}))\\
])~x&
\end{align*}

\textbf{Example 3}: AutoSWAP labeling function for the mouse behavior classification task. Since the mouse task is a multi-class classification task (for the ``attack'', ``investigate'', and ``mount'' behaviors), this program is not as trivially interpretable as those for the Fly and Basketball domains. Nevertheless, we can still deduce that the program classifies the angular domain-level LFs if the distance is small and the speed domain-level LFs otherwise. This is reasonable, as if the mice are far apart, the speed of the mice is probably sufficient for determining which behavior they are engaging in, and if they are close to each other, more detailed data (such as the angular LFs) may be needed.
\begin{align*}
\mapc(&\func~x_t~.~[\\
&\ifc~\textit{MouseDistanceAffine}(x)_{[1.02, -0.38]}\\
&\thenc~\textit{MouseAngularAffine}(x)\\
&\elsec~\textit{MouseSpeedAffine}(x)\\
])~x&
\end{align*}


%% file: supp_add_results.tex
\section{Additional Results}~\label{sec:add_results}
\begin{figure}[!h]
\includegraphics[width=0.33\linewidth]{diagrams/Random_Sampling_Fly.png}
\includegraphics[width=0.33\linewidth]{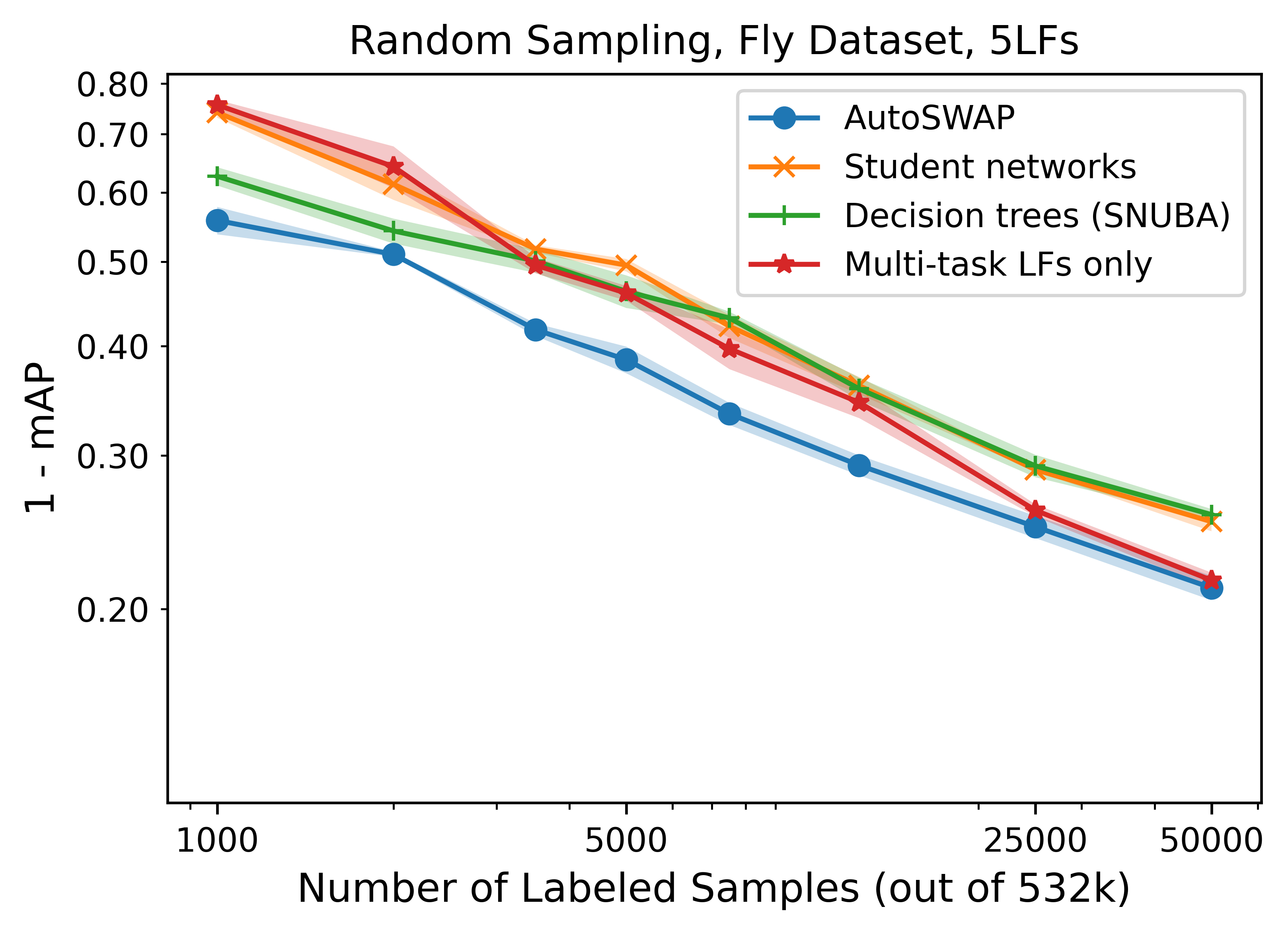}
\includegraphics[width=0.33\linewidth]{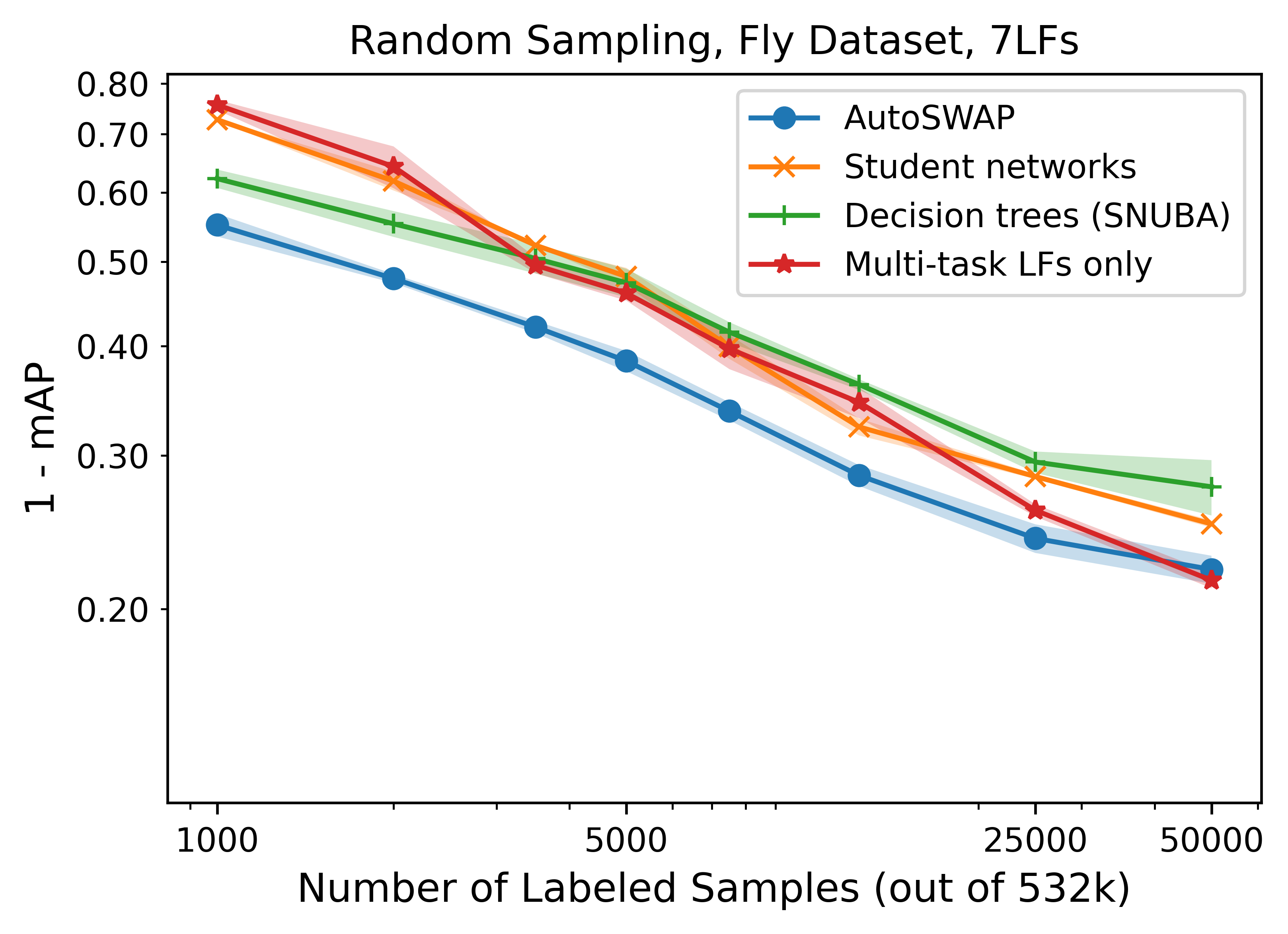}\\
\includegraphics[width=0.33\linewidth]{diagrams/Active_Learning_Fly.png}
\includegraphics[width=0.33\linewidth]{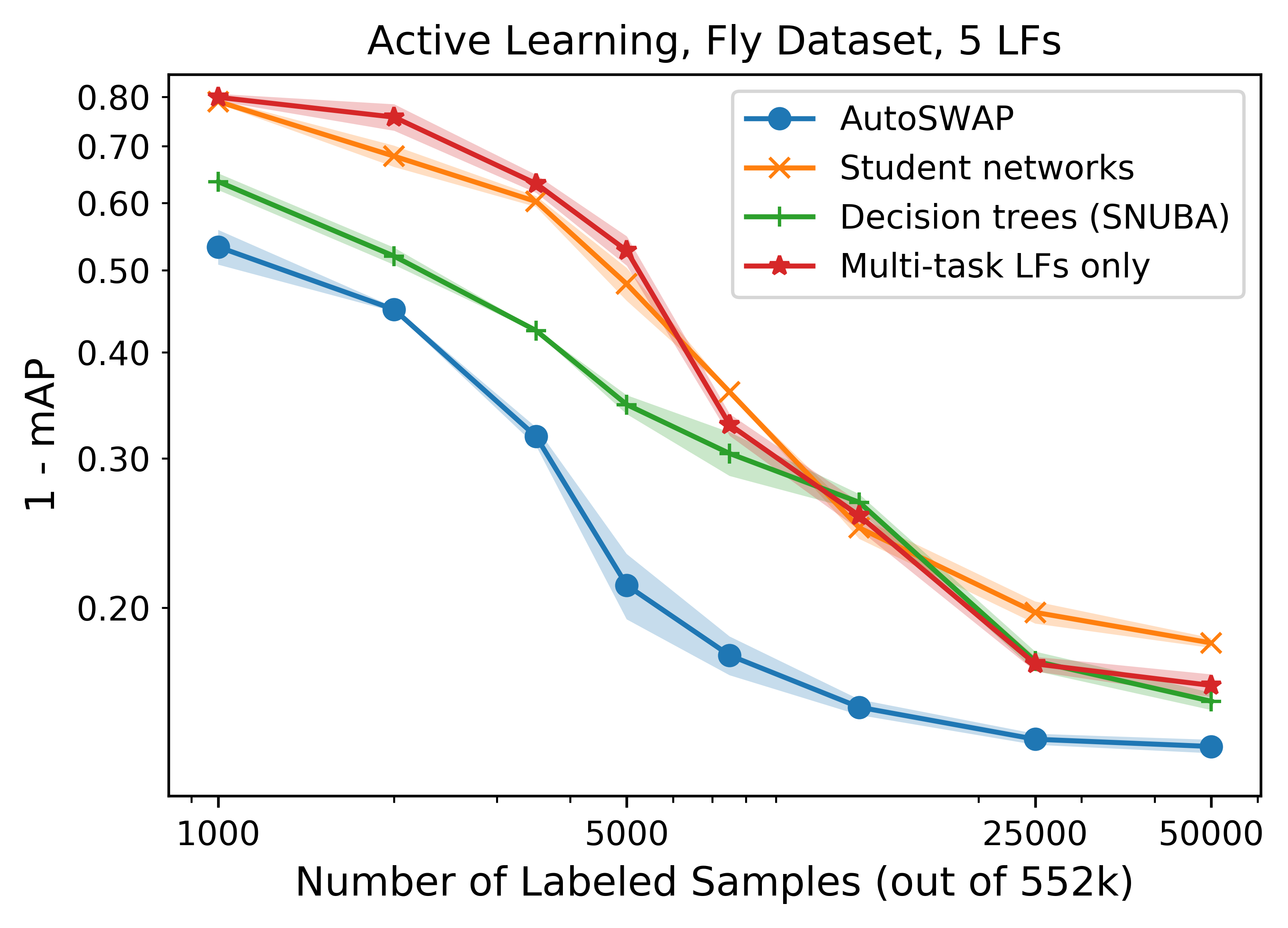}
\includegraphics[width=0.33\linewidth]{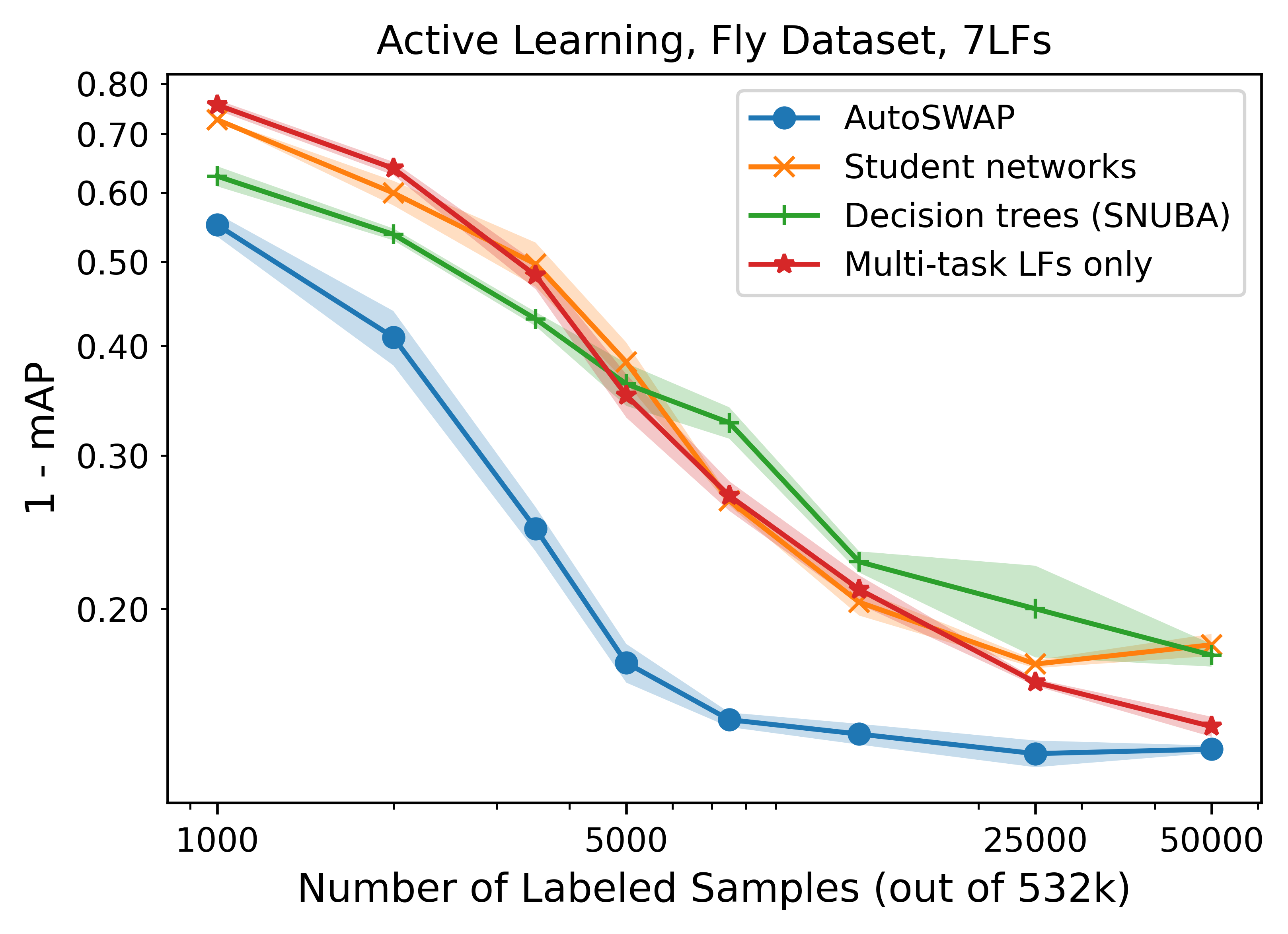}\\
\includegraphics[width=0.33\linewidth]{diagrams/Weak_Supervision_Fly.png}
\includegraphics[width=0.33\linewidth]{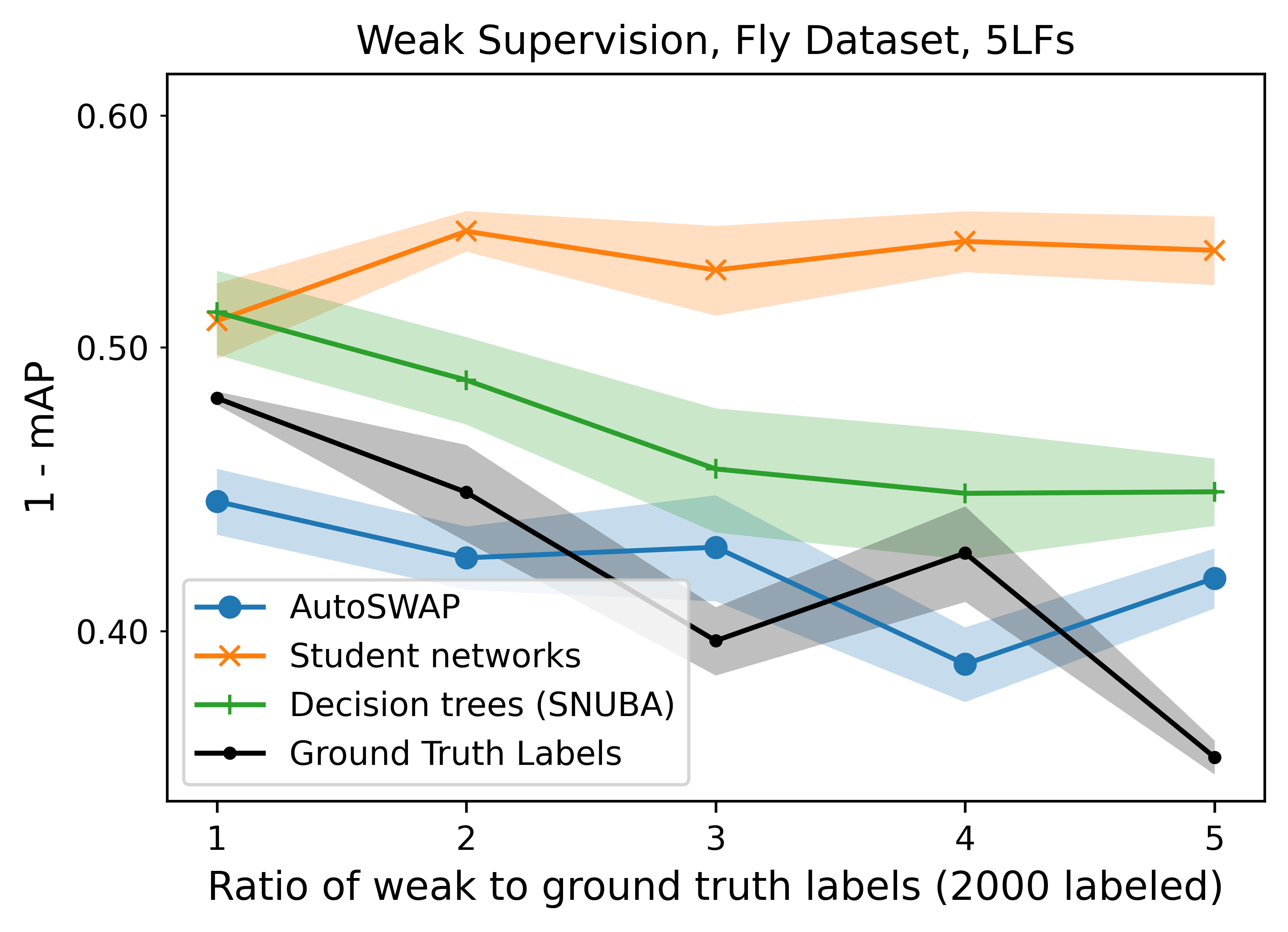}
\includegraphics[width=0.33\linewidth]{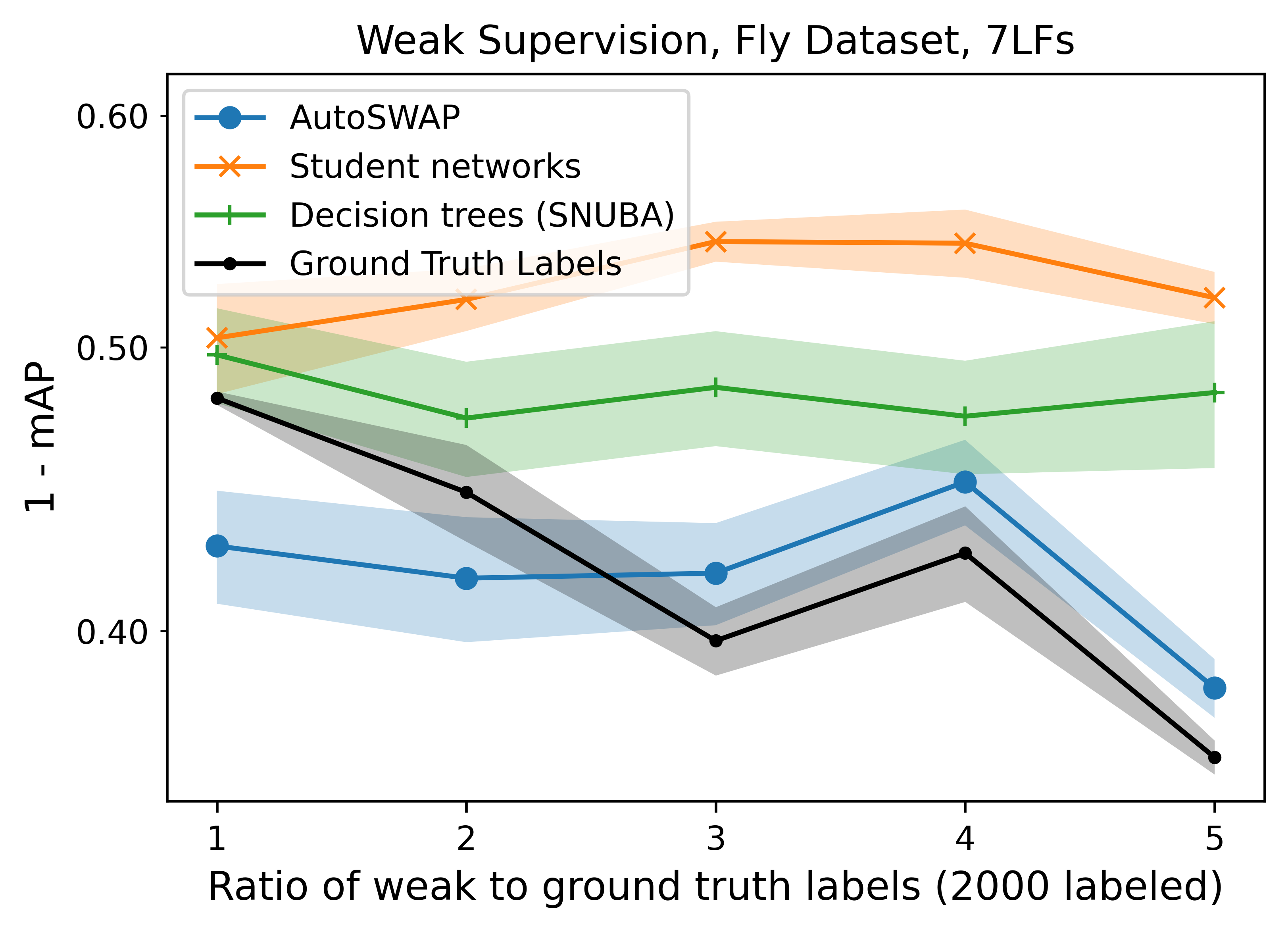}\\
\caption{Experiments with 3 (left-most column), 5, and 7 LFs for the Fly dataset. All plots are on a log-log scale. Due to resource constraints, some experiments were run on different machines, effectively giving different seeds. Results within each plot were run on the same machine and are directly comparable, but some results that should otherwise be the same across plots (e.g. ground truth labels) may have slight differences. We note that in all plots, regardless of seed, {\algname} outperforms the baselines.}
\label{fig:morelf_fly}
\end{figure}

\begin{figure}[!h]
\includegraphics[width=0.33\linewidth]{diagrams/Random_Sampling_Mouse.png}
\includegraphics[width=0.33\linewidth]{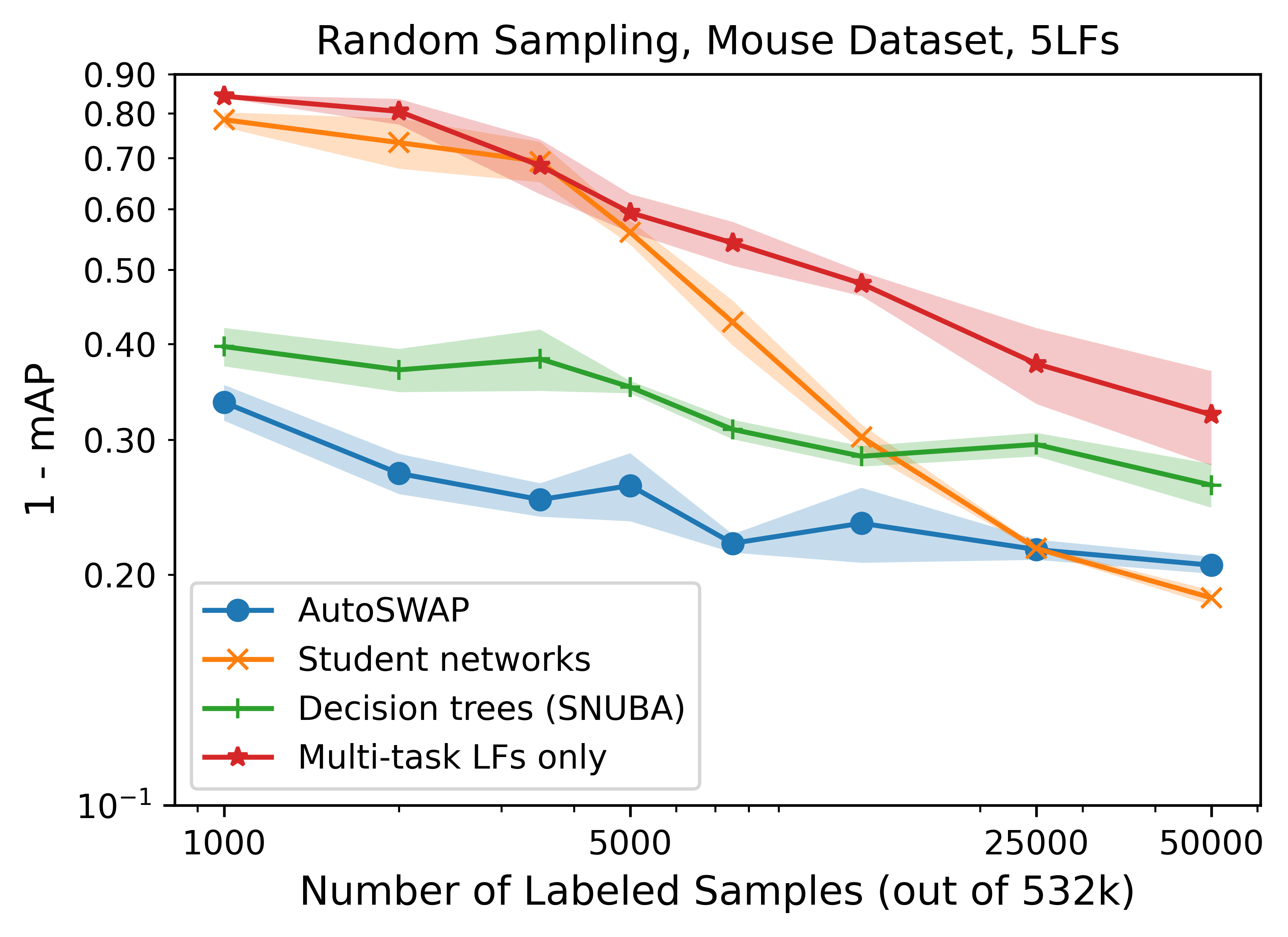}
\includegraphics[width=0.33\linewidth]{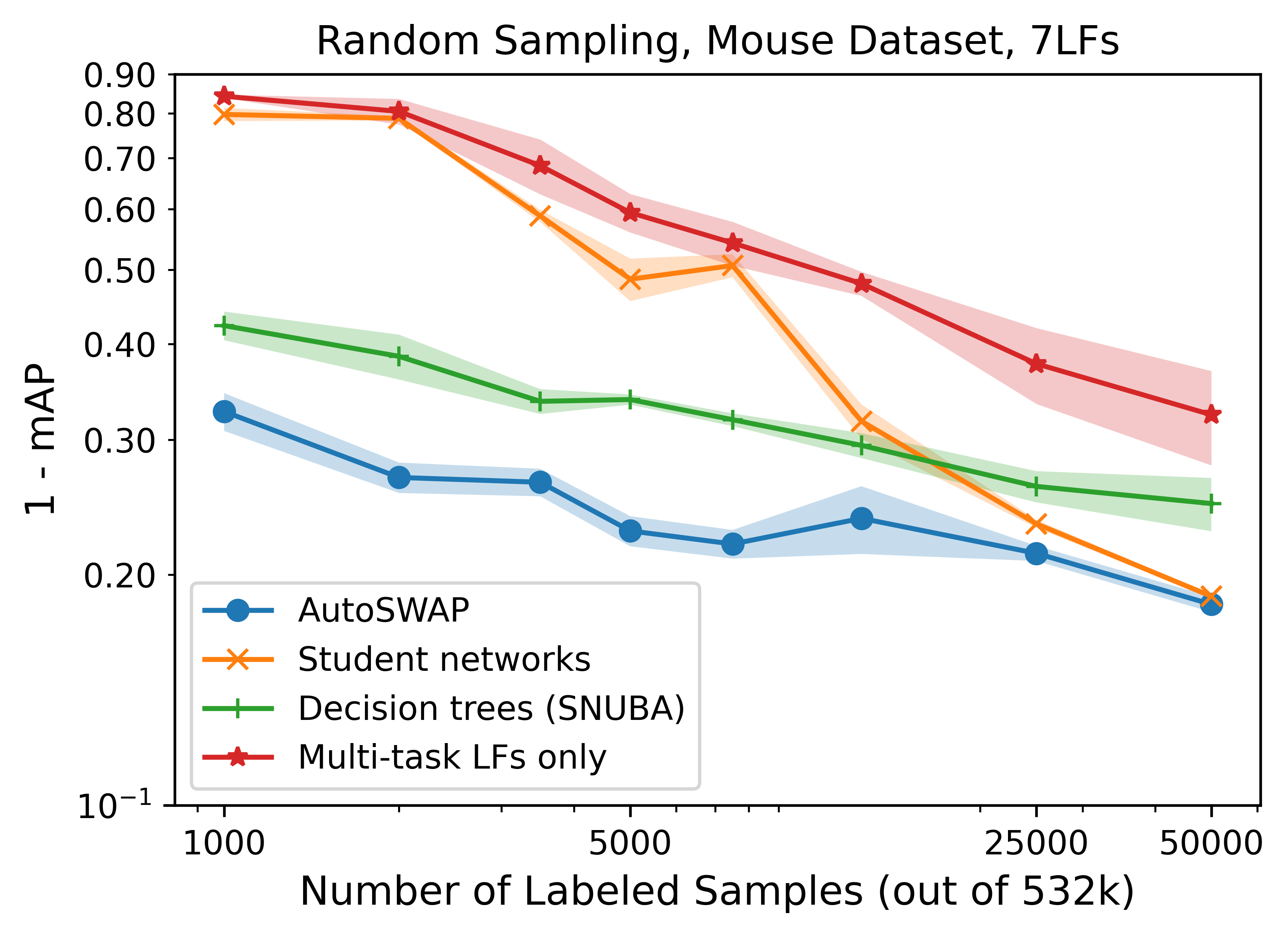}\\
\includegraphics[width=0.33\linewidth]{diagrams/Active_Learning_Mouse.png}
\includegraphics[width=0.33\linewidth]{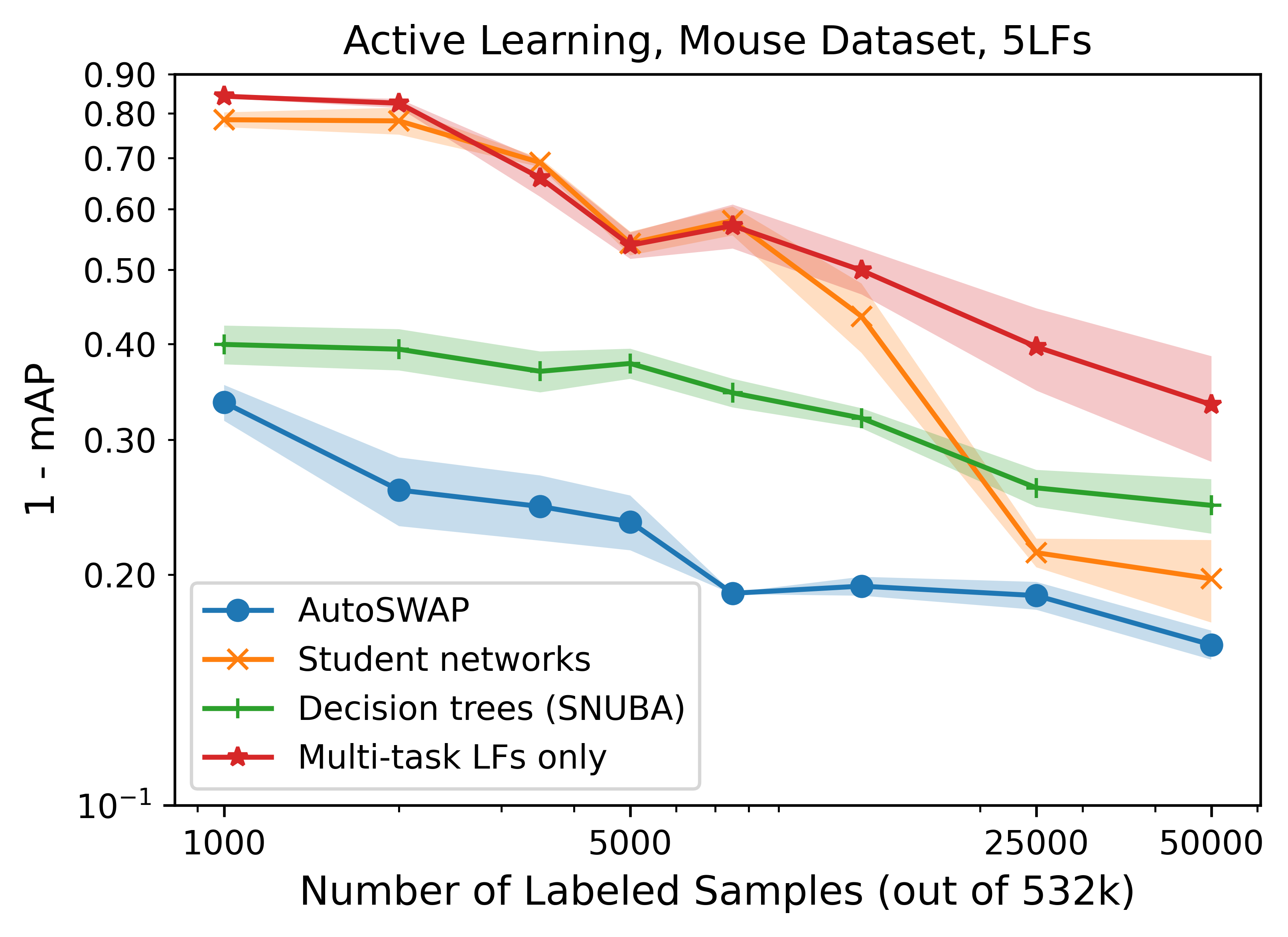}
\includegraphics[width=0.33\linewidth]{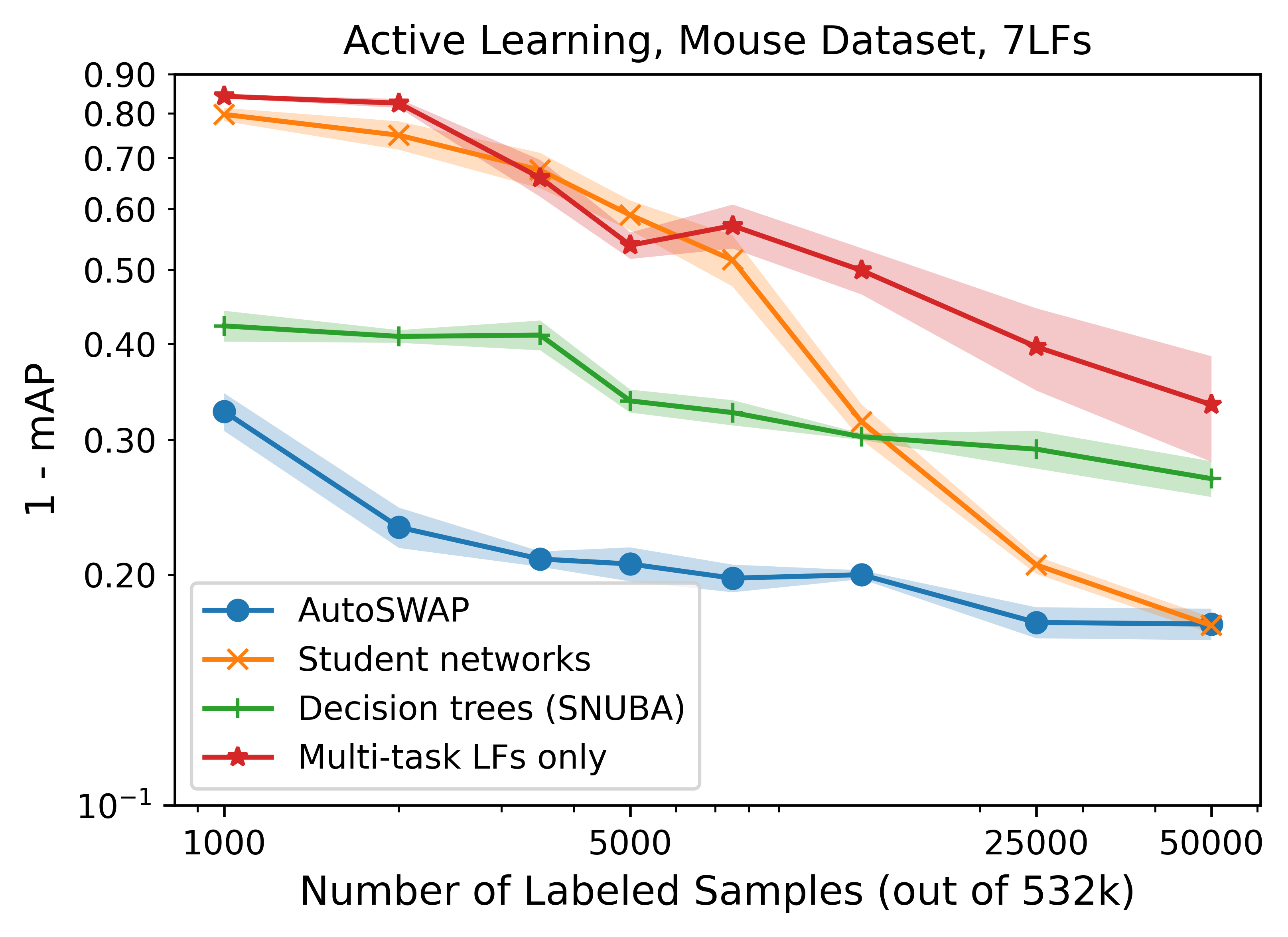}\\
\includegraphics[width=0.33\linewidth]{diagrams/Weak_Supervision_Mouse.png}
\includegraphics[width=0.33\linewidth]{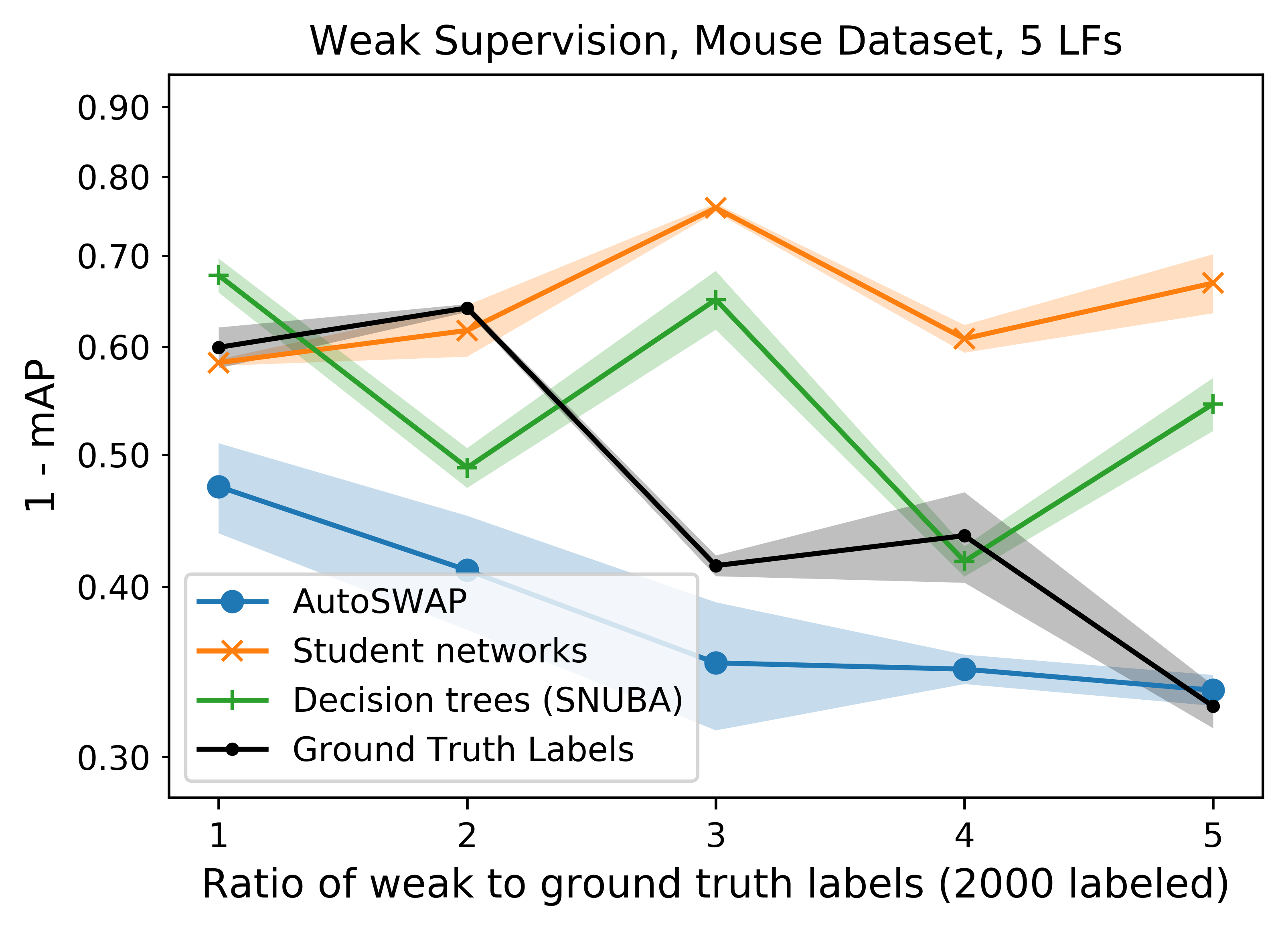}
\includegraphics[width=0.33\linewidth]{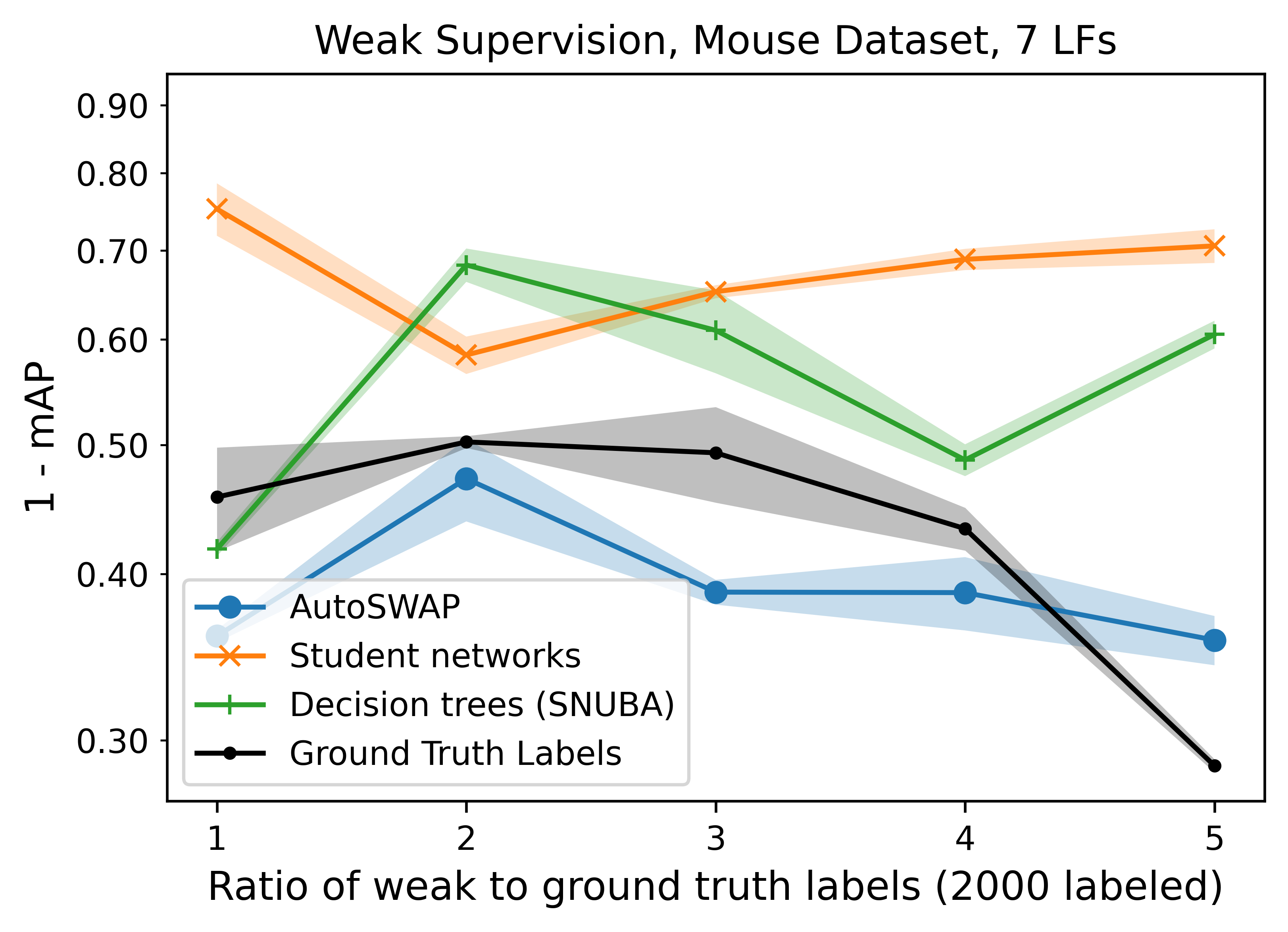}\\
\caption{Experiments with 3 (left-most column), 5, and 7 LFs for the Mouse dataset. All plots are on a log-log scale. Due to resource constraints, some experiments were run on different machines, effectively giving different seeds. Results within each plot were run on the same machine and are directly comparable, but some results that should otherwise be the same across plots (e.g. ground truth labels) may have slight differences. We note that in all plots, regardless of seed, {\algname} outperforms the baselines.}
\label{fig:morelf_mouse}
\end{figure}

\begin{figure}[!h]
\includegraphics[width=0.33\linewidth]{diagrams/Random_Sampling_Basketball.png}
\includegraphics[width=0.33\linewidth]{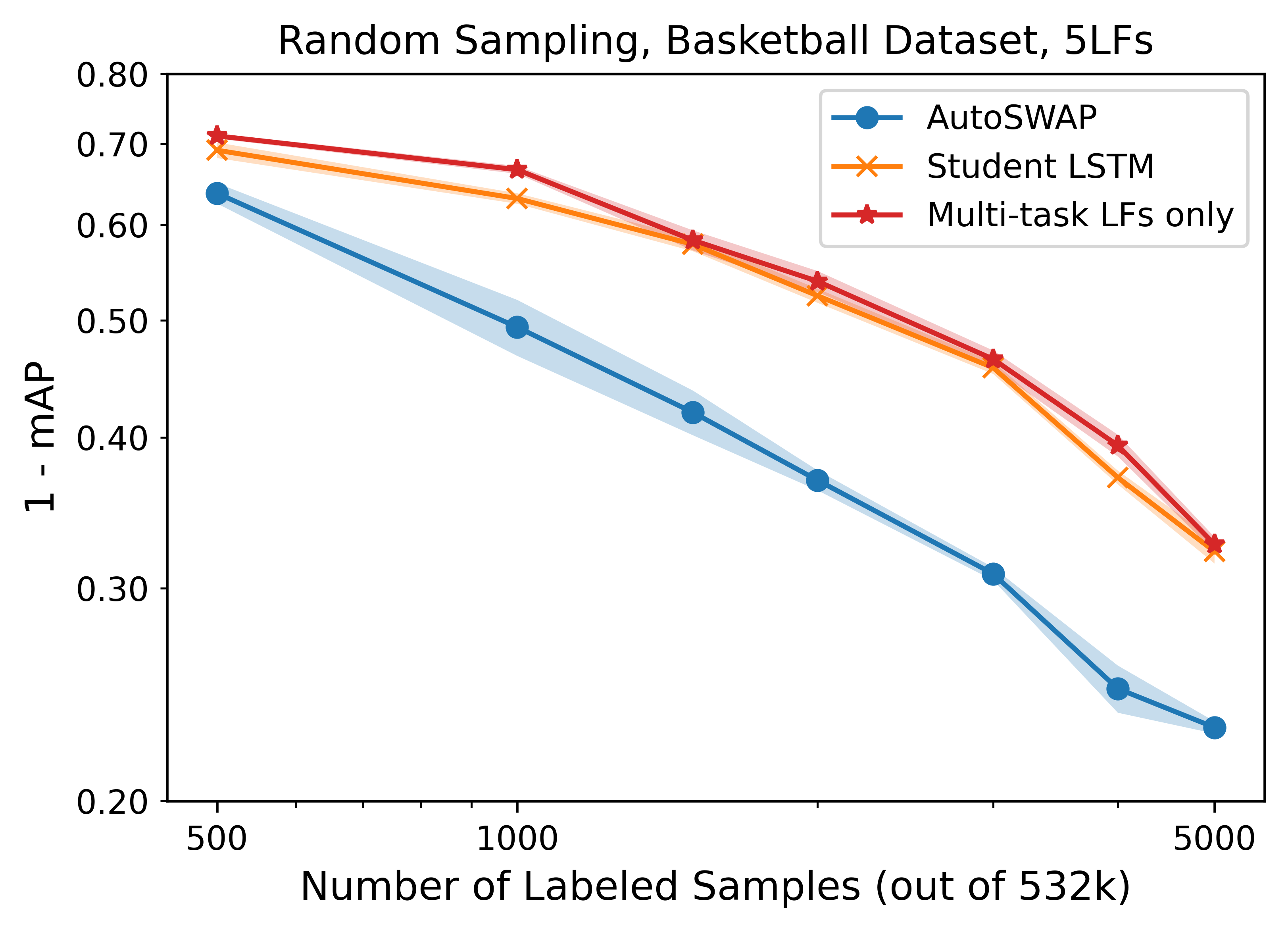}
\includegraphics[width=0.33\linewidth]{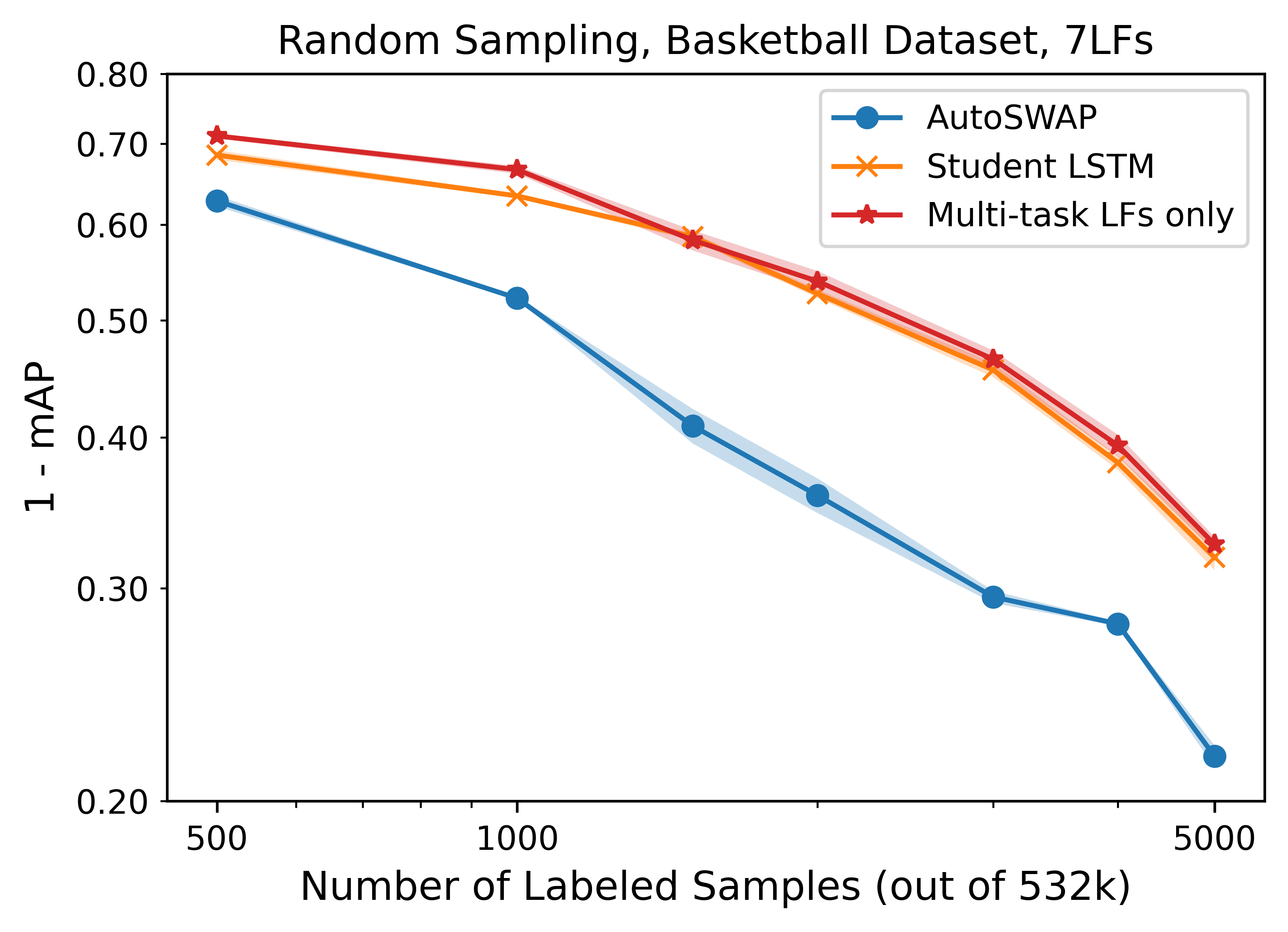}\\
\includegraphics[width=0.33\linewidth]{diagrams/Active_Learning_Basketball.png}
\includegraphics[width=0.33\linewidth]{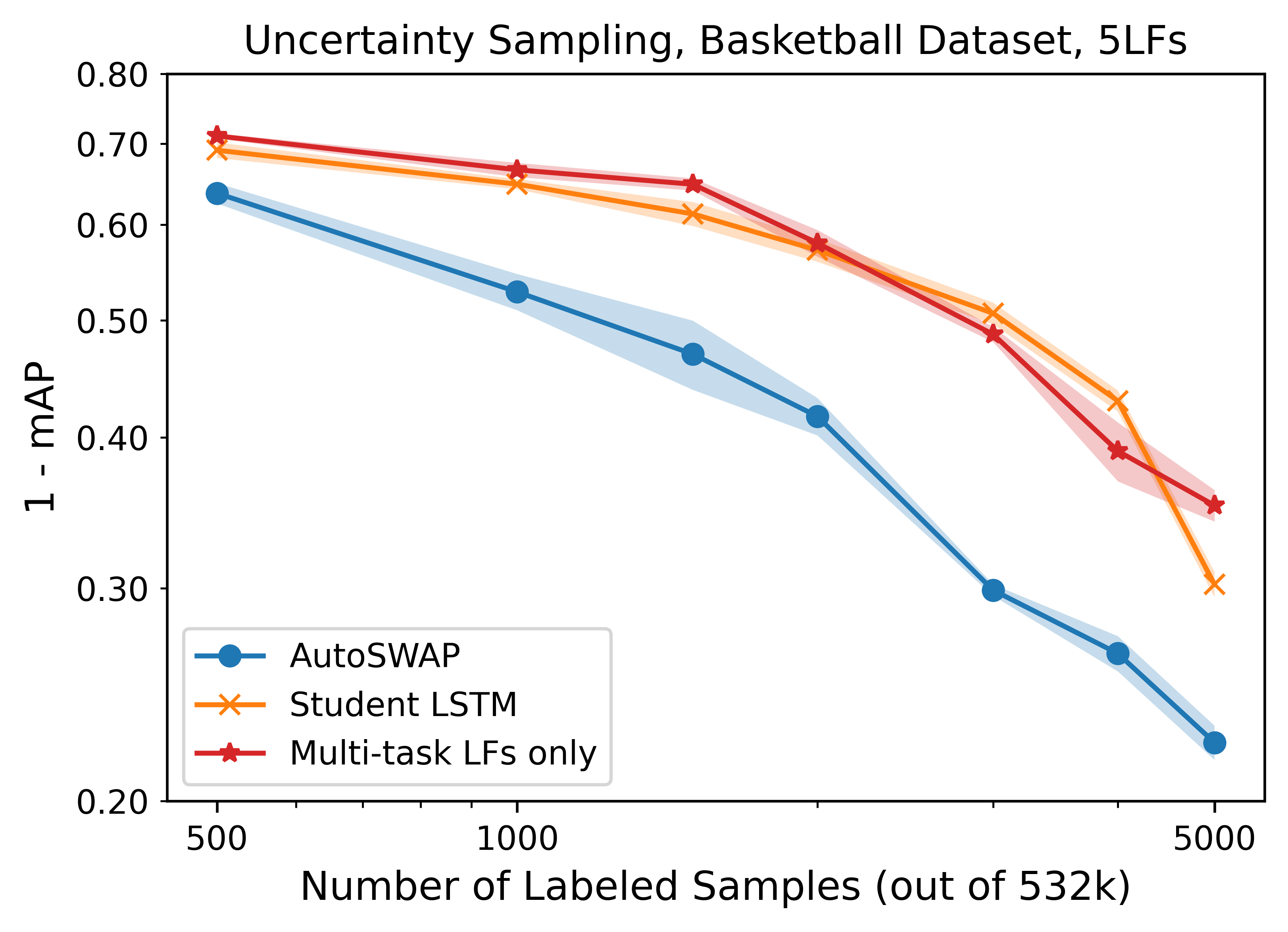}
\includegraphics[width=0.33\linewidth]{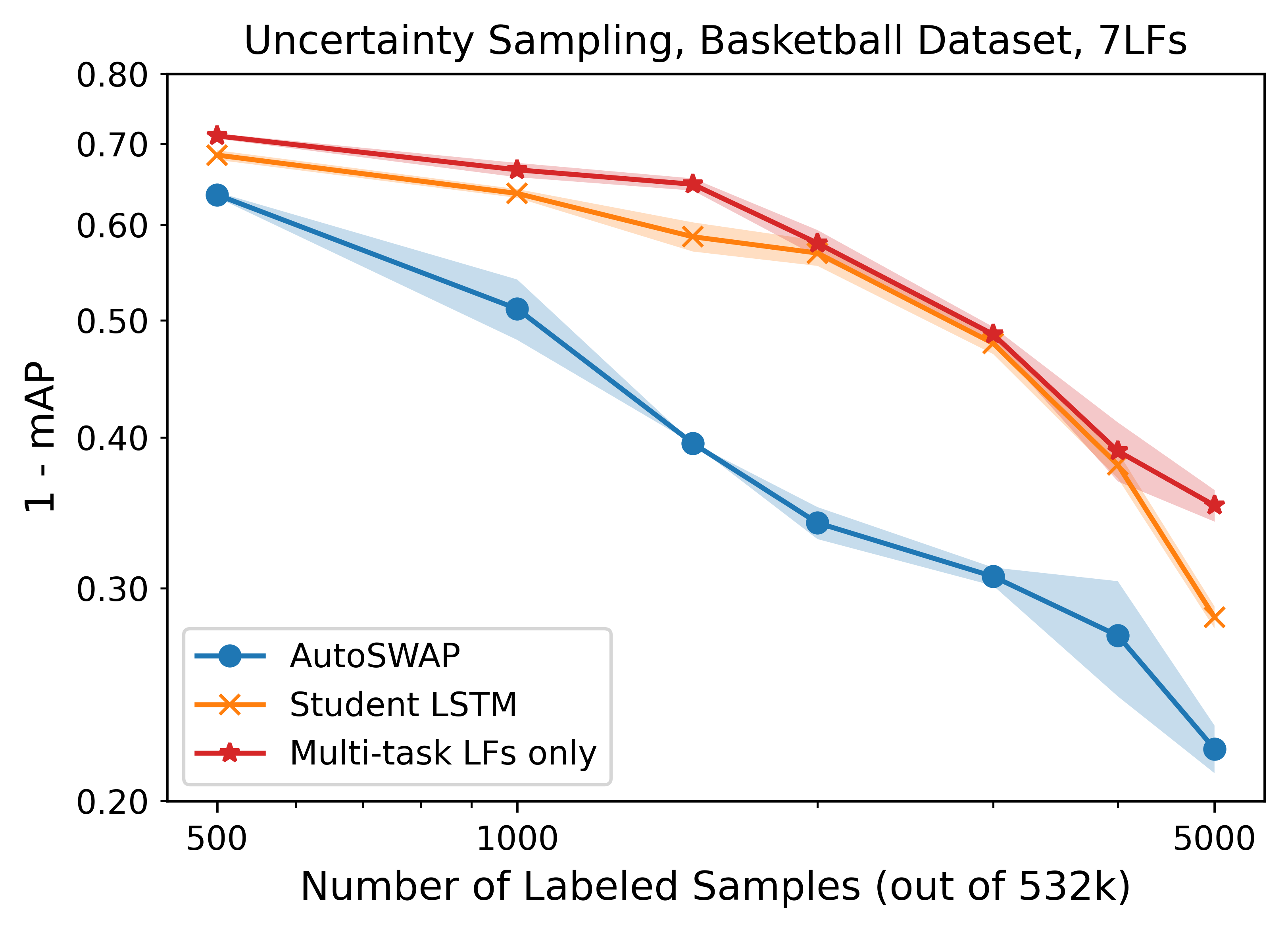}\\
\includegraphics[width=0.33\linewidth]{diagrams/Weak_Supervision_Basketball.png}
\includegraphics[width=0.33\linewidth]{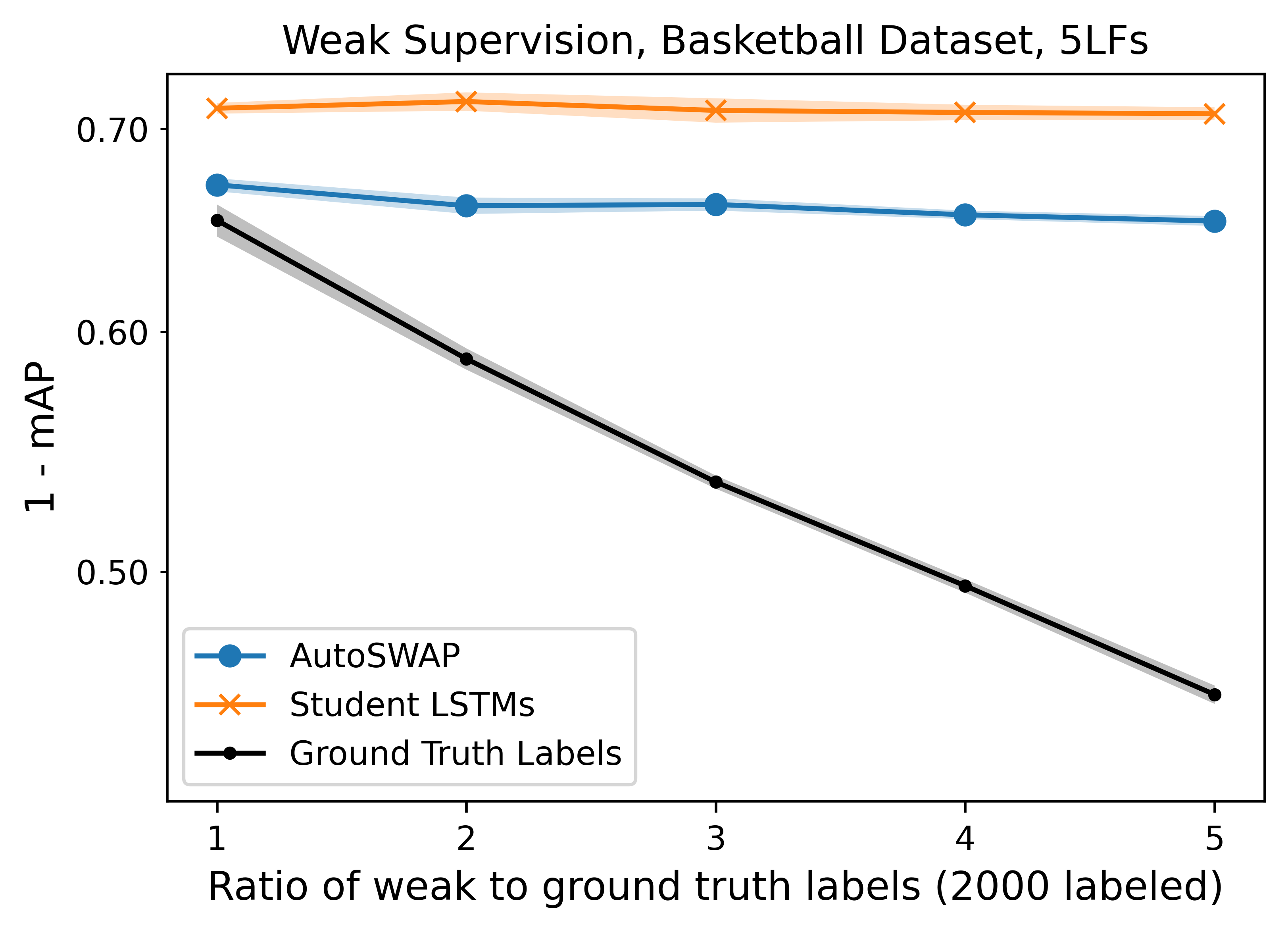}
\includegraphics[width=0.33\linewidth]{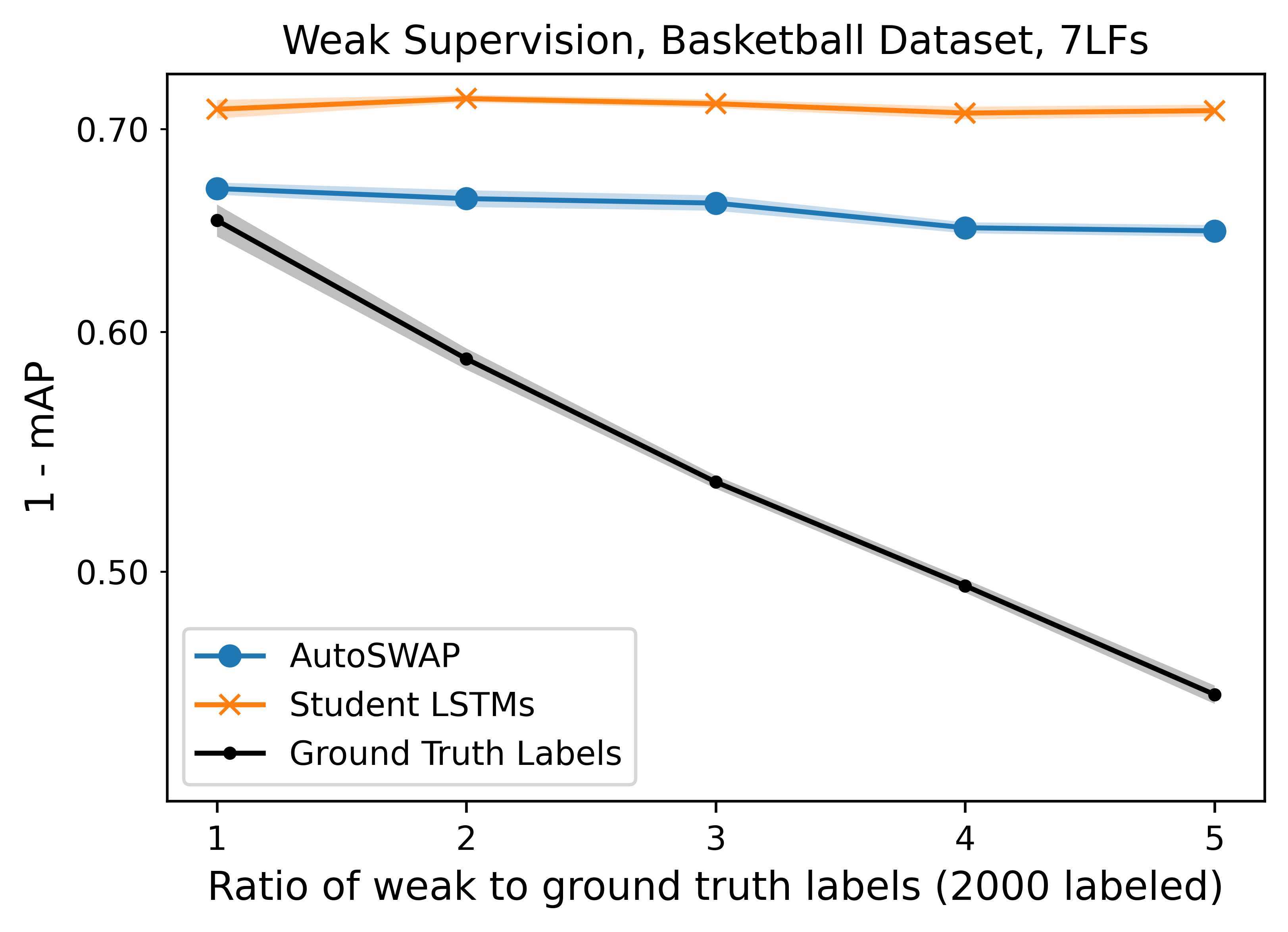}\\
\caption{Experiments with 3 (left-most column), 5, and 7 LFs for the Basketball dataset. All plots are on a log-log scale. Due to resource constraints, some experiments were run on different machines, effectively giving different seeds. Results within each plot were run on the same machine and are directly comparable, but some results that should otherwise be the same across plots (e.g. ground truth labels) may have slight differences. We note that in all plots, regardless of seed, {\algname} outperforms the baselines.}
\label{fig:morelf_bball}
\end{figure}

\textbf{Using More Labeling Functions}.
In our main experiments, we used 3 labeling functions.
The plots below (Figures \ref{fig:morelf_fly}, \ref{fig:morelf_mouse}, \ref{fig:morelf_bball}) show the effect of using more labeling functions (5, 7).
In general, active learning and random sampling performance stays relatively stable, but some improvement can be observed in weak supervision settings.
This can most likely be attributed to the generative weak label model performing better with more labeling functions, as well as improved coverage from the diversity measures.

\begin{figure}[!h]
\includegraphics[width=0.33\linewidth]{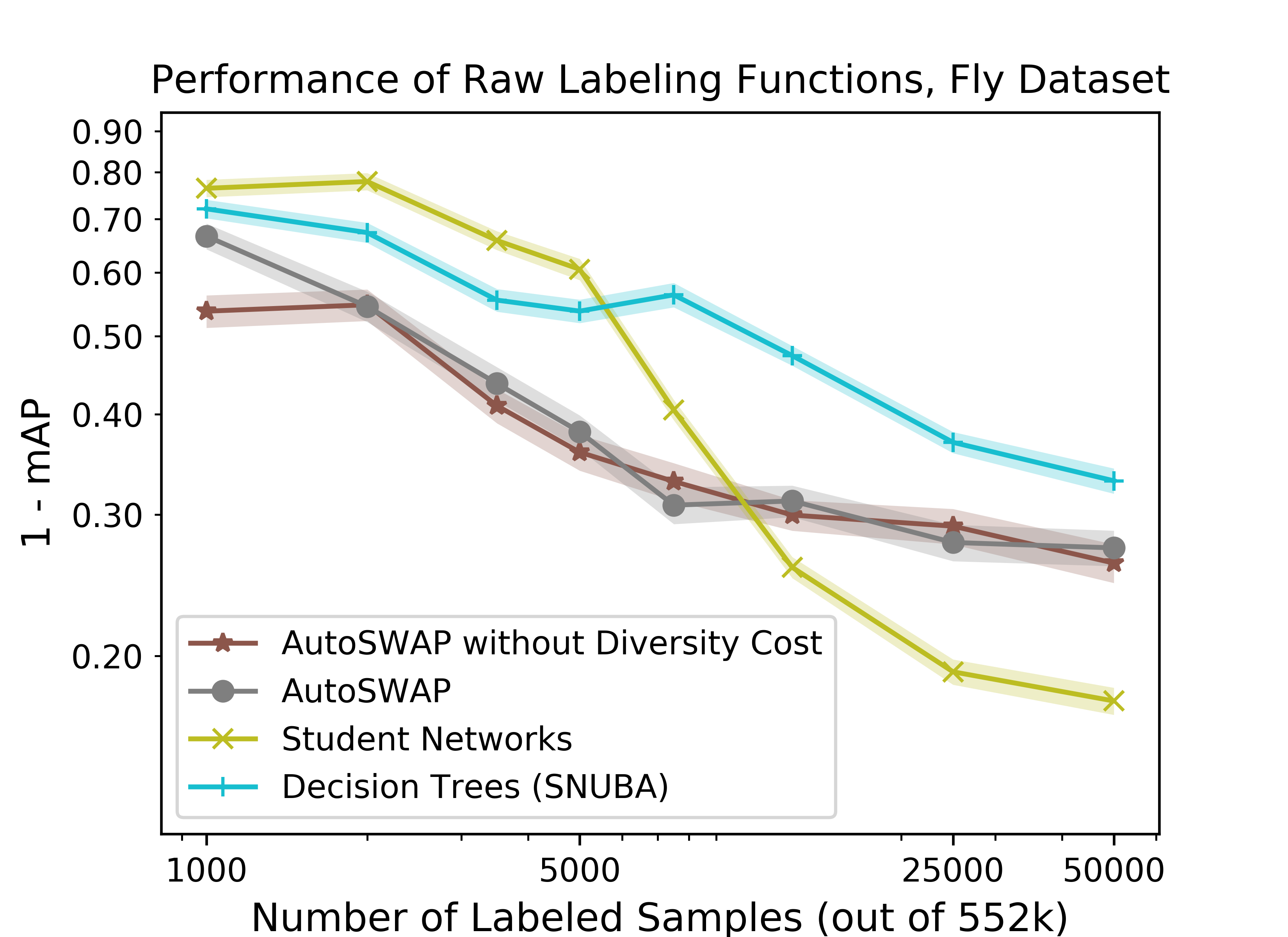}
\includegraphics[width=0.33\linewidth]{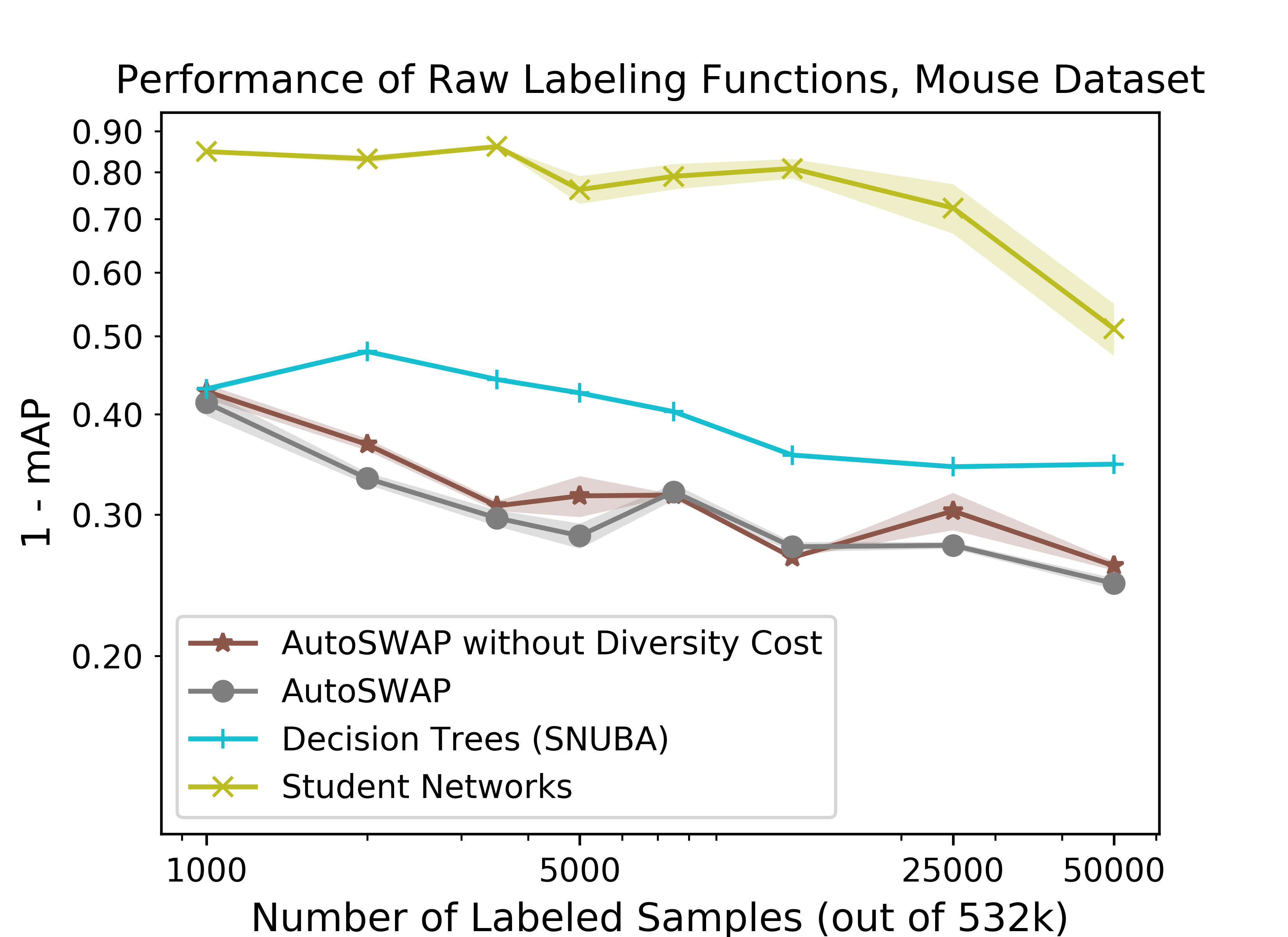}
\includegraphics[width=0.33\linewidth]{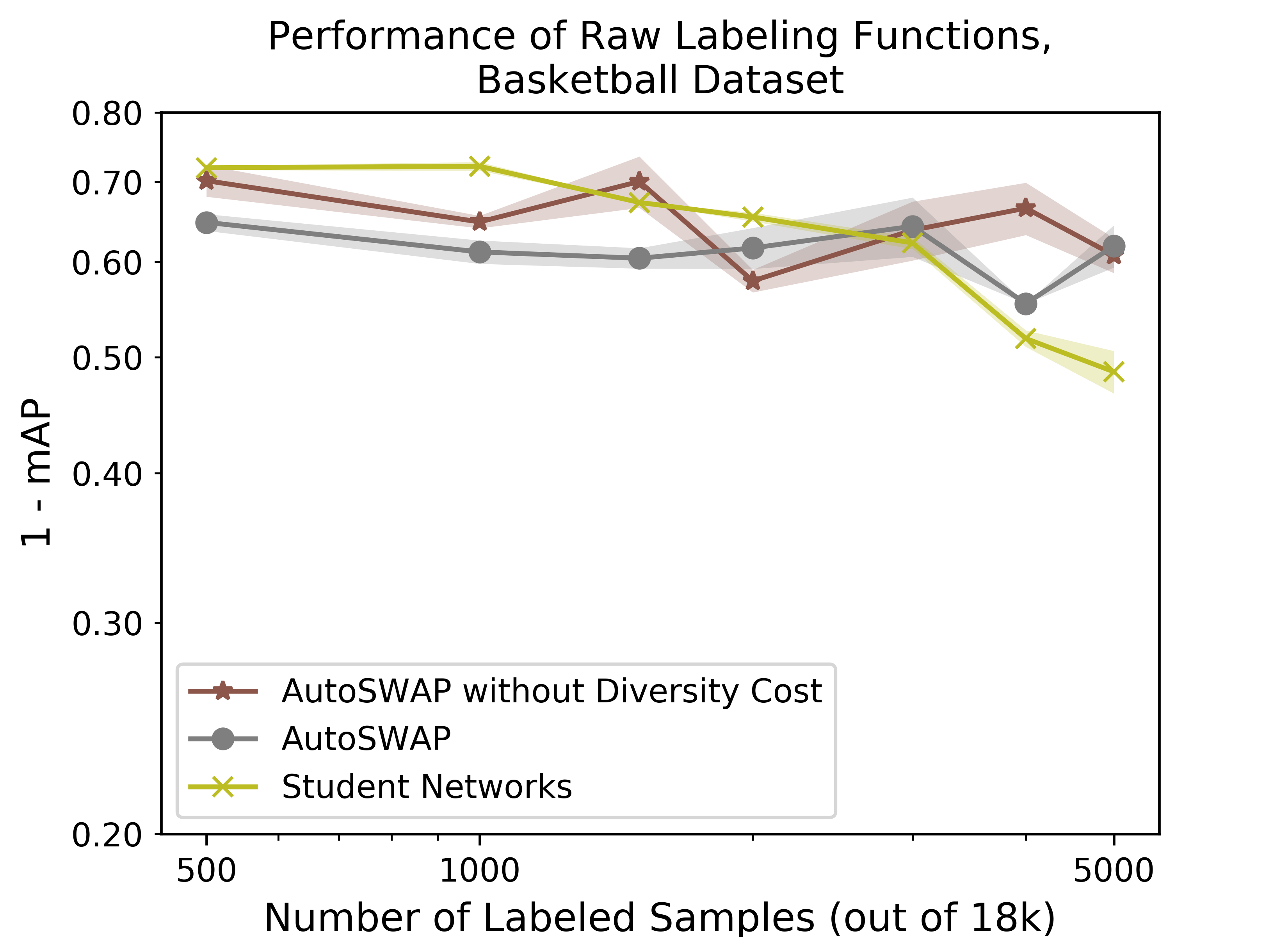}
\caption{Performance of generated task-level labeling functions outside of downstream tasks. As can be seen, AutoSWAP does not always outperform the baseline LF generation methods. This indicates that the data efficiency benefits of AutoSWAP come from the improved learning signal of AutoSWAP LFs relative to baseline LFs in downstream tasks. Furthermore, AutoSWAP LFs generated without the diversity cost do not perform significantly differently than those generated with the diversity cost, indicating that the diversity cost further improves the quality of the LFs' learning signal.}
\label{fig:rawlf}
\end{figure}

\FloatBarrier
\textbf{Isolated Labeling Function Performance}.
Here, we present the performance of the generated task-level labeling functions themselves, outside of any downstream tasks. Since task-level LFs give labels for the task at hand, they can be evaluated in the same context as the downstream tasks. As can be seen in Figure \ref{fig:rawlf}, AutoSWAP LFs do not always outperform LFs from student networks and decision trees. This indicates that the data efficiency benefits of AutoSWAP come from the higher quality learning signals AutoSWAP LFs give in downstream tasks, rather than their raw performance in a vacuum. Furthermore, AutoSWAP LFs generated without the diversity cost do not perform significantly differently than those generated with the diversity cost, indicating that the diversity cost further improves the quality of the LFs' learning signal.

\FloatBarrier
\textbf{Effect of Diversity Cost}
Table \ref{tab:diversity} shows a numerical comparison of the mean pairwise edit distance of 5 AutoSWAP LFs, taken over 5 seeds. Adding the diversity cost results in increased pairwise edit distance over all three behavior domains and increased performance in our experiments.

\begin{figure}[!h]
\centering
\includegraphics[width=0.7\linewidth]{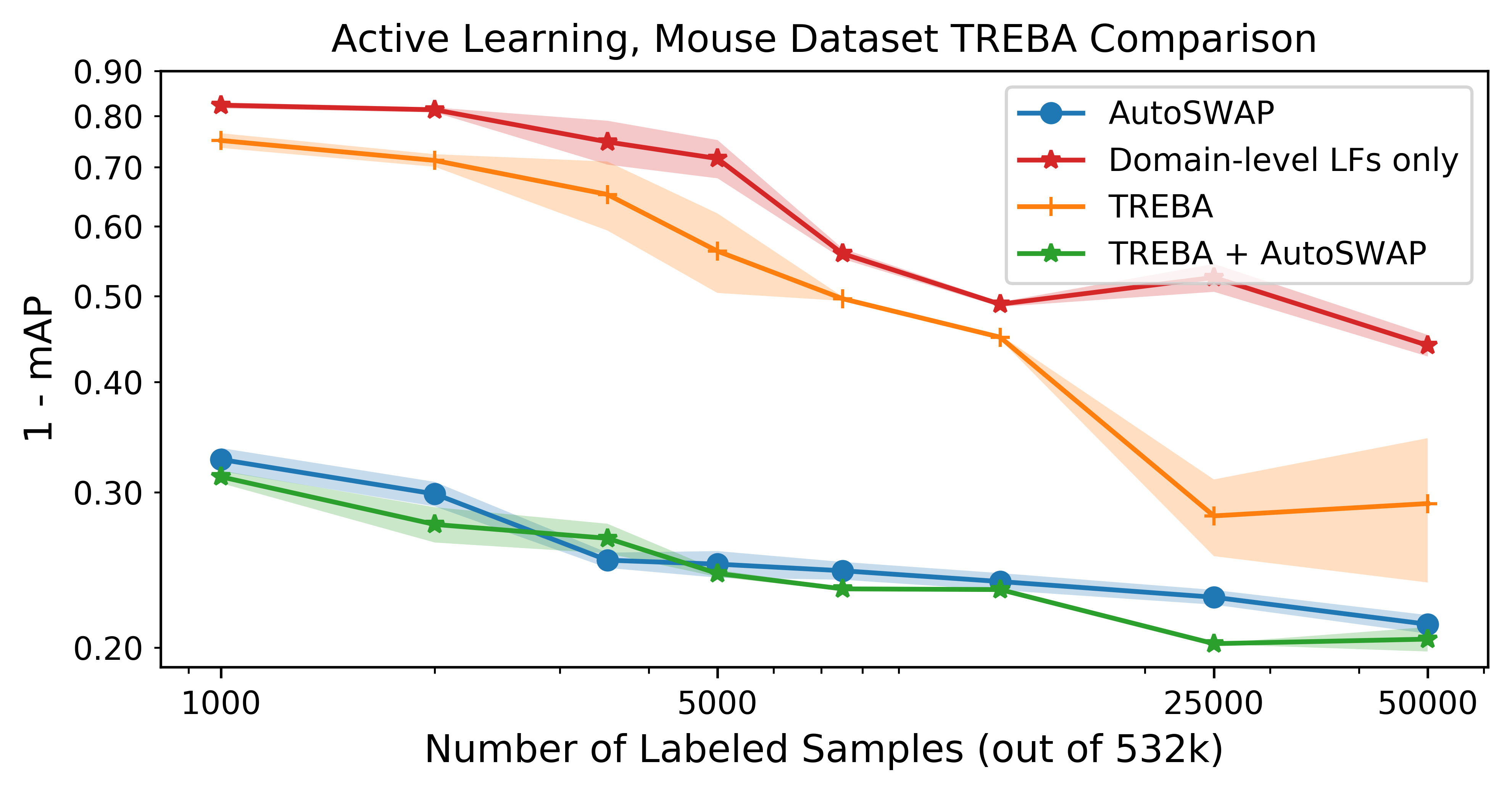}
\caption{Using self-supervised TREBA~\cite{sun2020task} features improves performance compared to only using domain-level LFs.}
\label{fig:treba}
\end{figure}

\begin{table}[!h]
\centering
\begin{tabular}{|l|ll|ll|ll|}
\hline
Frames & Fly         & No Diversity & Mouse       & No Diversity & Basketball       & No Diversity \\ \hline
2000   & \textbf{5.06} (0.08) & 4.08 (0.24)  & \textbf{4.62} (0.11) & 3.63 (0.29) & \textbf{3.36} (0.08) & 3.06 (0.13) \\
5000   & \textbf{5.00} (0.08) & 3.86 (0.14)  & \textbf{4.40} (0.22) & 4.17 (0.13) & \textbf{3.32} (0.09) & 3.08 (0.12)\\
12500  & \textbf{4.76} (0.04) & 3.50 (0.21)  & \textbf{4.58} (0.19) & 3.94 (0.14) & \textbf{3.24} (0.05) & 3.06 (0.06)\\
50000  & \textbf{4.90} (0.06) & 3.34 (0.39)  & \textbf{4.88} (0.12) & 3.77 (0.35) & \textbf{3.16} (0.04) & 2.86 (0.07)\\ \hline
\end{tabular}%
\caption{Mean pairwise edit distance of 5 AutoSWAP LFs with \& without diversity cost (5 seeds). Standard error in parentheses. Using the diversity cost results in programs that are ``farther apart'' and thus more structurally diverse.} \label{tab:diversity}
\end{table}
\FloatBarrier
\textbf{Compared to Self-Supervised Methods}
Self-supervised methods such as TREBA~\cite{sun2020task} have shown promise in reducing the number of labeled data points needed to achieve performance parity in behavior analysis domains.
Furthermore, such methods are generally complimentary to LF generation methods, as learned representations can be used as additional input features.
We provide an empirical analysis of combining AutoSWAP and TREBA in the Mouse domain active learning task in Figure \ref{fig:treba}, as TREBA features for CalMS21 are readily available.
We do not provide an analysis of TREBA for weak supervision task since TREBA features are \textit{features} and not \textit{labels}, and thus any comparisons between TREBA and AutoSWAP there would be indirect.
As can be seen in Figure \ref{fig:treba}, AutoSWAP alone outperforms TREBA, especially at lower levels of annotated samples, and using both gives slight performance benefits over using AutoSWAP alone.

\FloatBarrier

%% file: PaperForReview.bbl
\begin{thebibliography}{10}\itemsep=-1pt

\bibitem{anderson2014toward}
David~J Anderson and Pietro Perona.
\newblock Toward a science of computational ethology.
\newblock {\em Neuron}, 84(1):18--31, 2014.

\bibitem{ballhandler}
Jenna~Wiens Armand~McQueen and John Guttag.
\newblock Automatically recognizing on-ball screens.
\newblock In {\em MIT Sloan Sports Analytics Conference}, 2014.

\bibitem{berman2014mapping}
Gordon~J Berman, Daniel~M Choi, William Bialek, and Joshua~W Shaevitz.
\newblock Mapping the stereotyped behaviour of freely moving fruit flies.
\newblock {\em Journal of The Royal Society Interface}, 11(99):20140672, 2014.

\bibitem{biegel2021active}
Samantha Biegel, Rafah El-Khatib, Luiz Otavio Vilas~Boas Oliveira, Max Baak,
  and Nanne Aben.
\newblock Active weasul: Improving weak supervision with active learning.
\newblock {\em arXiv preprint arXiv:2104.14847}, 2021.

\bibitem{boecking2020interactive}
Benedikt Boecking, Willie Neiswanger, Eric Xing, and Artur Dubrawski.
\newblock Interactive weak supervision: Learning useful heuristics for data
  labeling.
\newblock In {\em International Conference on Learning Representations}, 2021.

\bibitem{calhoun2019unsupervised}
Adam~J Calhoun, Jonathan~W Pillow, and Mala Murthy.
\newblock Unsupervised identification of the internal states that shape natural
  behavior.
\newblock {\em Nature neuroscience}, 22(12):2040--2049, 2019.

\bibitem{chang2019argoverse}
Ming-Fang Chang, John Lambert, Patsorn Sangkloy, Jagjeet Singh, Slawomir Bak,
  Andrew Hartnett, De Wang, Peter Carr, Simon Lucey, Deva Ramanan, et~al.
\newblock Argoverse: 3d tracking and forecasting with rich maps.
\newblock In {\em Proceedings of the IEEE Conference on Computer Vision and
  Pattern Recognition}, pages 8748--8757, 2019.

\bibitem{chen2021web}
Qiaochu Chen, Aaron Lamoreaux, Xinyu Wang, Greg Durrett, Osbert Bastani, and
  Isil Dillig.
\newblock Web question answering with neurosymbolic program synthesis.
\newblock In {\em Proceedings of the 42nd ACM SIGPLAN International Conference
  on Programming Language Design and Implementation}, pages 328--343, 2021.

\bibitem{colyar2007us}
James Colyar and John Halkias.
\newblock Us highway 101 dataset.
\newblock {\em Federal Highway Administration (FHWA), Tech. Rep.
  FHWA-HRT-07-030}, 2007.

\bibitem{dunnmon2019}
Jared~A. Dunnmon, Alexander~J. Ratner, Khaled Saab, Nishith Khandwala, Matthew
  Markert, Hersh Sagreiya, Roger Goldman, Christopher Lee-Messer, Matthew~P.
  Lungren, Daniel~L. Rubin, and Christopher Ré.
\newblock Cross-modal data programming enables rapid medical machine learning.
\newblock {\em Patterns}, 1(2):100019, 2020.

\bibitem{NEURIPS2018_7aa685b3}
Kevin Ellis, Lucas Morales, Mathias Sabl\'{e}-Meyer, Armando Solar-Lezama, and
  Josh Tenenbaum.
\newblock Learning libraries of subroutines for neurally\textendash guided
  bayesian program induction.
\newblock In S. Bengio, H. Wallach, H. Larochelle, K. Grauman, N. Cesa-Bianchi,
  and R. Garnett, editors, {\em Advances in Neural Information Processing
  Systems}, volume~31. Curran Associates, Inc., 2018.

\bibitem{ellis2017learning}
Kevin Ellis, Daniel Ritchie, Armando Solar-Lezama, and Josh Tenenbaum.
\newblock Learning to infer graphics programs from hand-drawn images.
\newblock In S. Bengio, H. Wallach, H. Larochelle, K. Grauman, N. Cesa-Bianchi,
  and R. Garnett, editors, {\em Advances in Neural Information Processing
  Systems}, volume~31. Curran Associates, Inc., 2018.

\bibitem{ellis2015unsupervised}
Kevin Ellis, Armando Solar-Lezama, and Josh Tenenbaum.
\newblock Unsupervised learning by program synthesis.
\newblock 2015.

\bibitem{eyjolfsdottir2014detecting}
Eyrun Eyjolfsdottir, Steve Branson, Xavier~P Burgos-Artizzu, Eric~D Hoopfer,
  Jonathan Schor, David~J Anderson, and Pietro Perona.
\newblock Detecting social actions of fruit flies.
\newblock In {\em European Conference on Computer Vision}, pages 772--787.
  Springer, 2014.

\bibitem{feser2015synthesizing}
John~K Feser, Swarat Chaudhuri, and Isil Dillig.
\newblock Synthesizing data structure transformations from input-output
  examples.
\newblock {\em ACM SIGPLAN Notices}, 50(6):229--239, 2015.

\bibitem{harris}
Larry~R. Harris.
\newblock The heuristic search under conditions of error.
\newblock {\em Artif. Intell.}, 5:217--234, 1974.

\bibitem{karamanolakis2021self}
Giannis Karamanolakis, Subhabrata~(Subho) Mukherjee, Guoqing Zheng, and
  Ahmed~H. Awadallah.
\newblock Self-training with weak supervision.
\newblock In {\em NAACL 2021}. NAACL 2021, May 2021.

\bibitem{lewis1996uncertainty}
David Lewis, Jason Catlett, W. Cohen, and Haym Hirsh.
\newblock Heterogeneous uncertainty sampling for supervised learning.
\newblock 12 1996.

\bibitem{luxem2020identifying}
Kevin Luxem, Falko Fuhrmann, Johannes K{\"u}rsch, Stefan Remy, and Pavol Bauer.
\newblock Identifying behavioral structure from deep variational embeddings of
  animal motion.
\newblock {\em bioRxiv}, 2020.

\bibitem{nashaat2018hybridization}
Mona Nashaat, Aindrila Ghosh, James Miller, Shaikh Quader, Chad Marston, and
  Jean-Francois Puget.
\newblock Hybridization of active learning and data programming for labeling
  large industrial datasets.
\newblock In {\em 2018 IEEE International Conference on Big Data (Big Data)},
  pages 46--55. IEEE, 2018.

\bibitem{parisotto2016neuro}
Emilio Parisotto, Abdel-rahman Mohamed, Rishabh Singh, Lihong Li, Dengyong
  Zhou, and Pushmeet Kohli.
\newblock Neuro-symbolic program synthesis.
\newblock {\em arXiv preprint arXiv:1611.01855}, 2016.

\bibitem{ratner_metal}
Alexander Ratner, Braden Hancock, Jared Dunnmon, Frederic Sala, Shreyash
  Pandey, and Christopher Ré.
\newblock Training complex models with multi-task weak supervision.
\newblock {\em Proceedings of the AAAI Conference on Artificial Intelligence},
  33(01):4763--4771, Jul. 2019.

\bibitem{ratner2016data}
Alexander~J Ratner, Christopher~M De~Sa, Sen Wu, Daniel Selsam, and Christopher
  R{\'e}.
\newblock Data programming: Creating large training sets, quickly.
\newblock {\em Advances in neural information processing systems},
  29:3567--3575, 2016.

\bibitem{segalin2020mouse}
Cristina Segalin, Jalani Williams, Tomomi Karigo, May Hui, Moriel Zelikowsky,
  Jennifer~J. Sun, Pietro Perona, David~J. Anderson, and Ann Kennedy.
\newblock The mouse action recognition system (mars): a software pipeline for
  automated analysis of social behaviors in mice.
\newblock {\em bioRxiv https://doi.org/10.1101/2020.07.26.222299}, 2020.

\bibitem{shah2020learning}
Ameesh Shah, Eric Zhan, Jennifer~J Sun, Abhinav Verma, Yisong Yue, and Swarat
  Chaudhuri.
\newblock Learning differentiable programs with admissible neural heuristics.
\newblock In {\em Neural Information Processing Systems}, 2020.

\bibitem{solar2006combinatorial}
Armando Solar-Lezama, Liviu Tancau, Rastislav Bodik, Sanjit Seshia, and Vijay
  Saraswat.
\newblock Combinatorial sketching for finite programs.
\newblock In {\em Proceedings of the 12th international conference on
  Architectural support for programming languages and operating systems}, pages
  404--415, 2006.

\bibitem{sun2021multi}
Jennifer~J. Sun, Tomomi Karigo, Dipam Chakraborty, Sharada Mohanty, Benjamin
  Wild, Quan Sun, Chen Chen, David Anderson, Pietro Perona, Yisong Yue, and Ann
  Kennedy.
\newblock The multi-agent behavior dataset: Mouse dyadic social interactions.
\newblock In {\em Thirty-fifth Conference on Neural Information Processing
  Systems Datasets and Benchmarks Track (Round 1)}, 2021.

\bibitem{sun2020task}
Jennifer~J Sun, Ann Kennedy, Eric Zhan, David~J Anderson, Yisong Yue, and
  Pietro Perona.
\newblock Task programming: Learning data efficient behavior representations.
\newblock In {\em Conference on Computer Vision and Pattern Recognition}, 2021.

\bibitem{sun2018dt}
Tao Sun and Zhi-Hua Zhou.
\newblock Structural diversity for decision tree ensemble learning.
\newblock {\em Frontiers of Computer Science}, 12, 02 2018.

\bibitem{tuyls2021game}
Karl Tuyls, Shayegan Omidshafiei, Paul Muller, Zhe Wang, Jerome Connor, Daniel
  Hennes, Ian Graham, William Spearman, Tim Waskett, Dafydd Steel, et~al.
\newblock Game plan: What ai can do for football, and what football can do for
  ai.
\newblock {\em Journal of Artificial Intelligence Research}, 71:41--88, 2021.

\bibitem{valkov2018houdini}
Lazar Valkov, Dipak Chaudhari, Akash Srivastava, Charles Sutton, and Swarat
  Chaudhuri.
\newblock Houdini: Lifelong learning as program synthesis.
\newblock In {\em Advances in neural information processing systems}, 2018.

\bibitem{NEURIPS2018_edc27f13}
Lazar Valkov, Dipak Chaudhari, Akash Srivastava, Charles Sutton, and Swarat
  Chaudhuri.
\newblock Houdini: Lifelong learning as program synthesis.
\newblock In S. Bengio, H. Wallach, H. Larochelle, K. Grauman, N. Cesa-Bianchi,
  and R. Garnett, editors, {\em Advances in Neural Information Processing
  Systems}, volume~31. Curran Associates, Inc., 2018.

\bibitem{varma2018snuba}
Paroma Varma and Christopher R{\'e}.
\newblock Snuba: Automating weak supervision to label training data.
\newblock In {\em Proceedings of the VLDB Endowment. International Conference
  on Very Large Data Bases}, volume~12, page 223. NIH Public Access, 2018.

\bibitem{verma2018programmatically}
Abhinav Verma, Vijayaraghavan Murali, Rishabh Singh, Pushmeet Kohli, and Swarat
  Chaudhuri.
\newblock Programmatically interpretable reinforcement learning.
\newblock In {\em International Conference on Machine Learning}, pages
  5045--5054. PMLR, 2018.

\bibitem{noisystudent}
Qizhe Xie, Minh-Thang Luong, Eduard Hovy, and Quoc~V. Le.
\newblock Self-training with noisy student improves imagenet classification,
  2020.

\bibitem{yue2014learning}
Yisong Yue, Patrick Lucey, Peter Carr, Alina Bialkowski, and Iain Matthews.
\newblock Learning fine-grained spatial models for dynamic sports play
  prediction.
\newblock In {\em 2014 IEEE international conference on data mining}, pages
  670--679. IEEE, 2014.

\bibitem{zhan2018generating}
Eric Zhan, Stephan Zheng, Yisong Yue, Long Sha, and Patrick Lucey.
\newblock Generating multi-agent trajectories using programmatic weak
  supervision.
\newblock In {\em International Conference on Learning Representations}, 2019.

\bibitem{zss}
Kaizhong Zhang and Dennis Shasha.
\newblock Simple fast algorithms for the editing distance between trees and
  related problems.
\newblock {\em SIAM J. Comput.}, 18:1245--1262, 12 1989.

\end{thebibliography}


\begin{thebibliography}{1}\itemsep=-1pt

\bibitem{ballhandler}
Jenna~Wiens Armand~McQueen and John Guttag.
\newblock Automatically recognizing on-ball screens.
\newblock In {\em MIT Sloan Sports Analytics Conference}, 2014.

\bibitem{eyjolfsdottir2014detecting}
Eyrun Eyjolfsdottir, Steve Branson, Xavier~P Burgos-Artizzu, Eric~D Hoopfer,
  Jonathan Schor, David~J Anderson, and Pietro Perona.
\newblock Detecting social actions of fruit flies.
\newblock In {\em European Conference on Computer Vision}, pages 772--787.
  Springer, 2014.

\bibitem{segalin2020mouse}
Cristina Segalin, Jalani Williams, Tomomi Karigo, May Hui, Moriel Zelikowsky,
  Jennifer~J. Sun, Pietro Perona, David~J. Anderson, and Ann Kennedy.
\newblock The mouse action recognition system (mars): a software pipeline for
  automated analysis of social behaviors in mice.
\newblock {\em bioRxiv https://doi.org/10.1101/2020.07.26.222299}, 2020.

\bibitem{sun2021multi}
Jennifer~J. Sun, Tomomi Karigo, Dipam Chakraborty, Sharada Mohanty, Benjamin
  Wild, Quan Sun, Chen Chen, David Anderson, Pietro Perona, Yisong Yue, and Ann
  Kennedy.
\newblock The multi-agent behavior dataset: Mouse dyadic social interactions.
\newblock In {\em Thirty-fifth Conference on Neural Information Processing
  Systems Datasets and Benchmarks Track (Round 1)}, 2021.

\bibitem{sun2020task}
Jennifer~J Sun, Ann Kennedy, Eric Zhan, David~J Anderson, Yisong Yue, and
  Pietro Perona.
\newblock Task programming: Learning data efficient behavior representations.
\newblock In {\em Conference on Computer Vision and Pattern Recognition}, 2021.

\bibitem{yue2014learning}
Yisong Yue, Patrick Lucey, Peter Carr, Alina Bialkowski, and Iain Matthews.
\newblock Learning fine-grained spatial models for dynamic sports play
  prediction.
\newblock In {\em 2014 IEEE international conference on data mining}, pages
  670--679. IEEE, 2014.

\end{thebibliography}
